\documentclass[sn-mathphys]{sn-jnl}

\usepackage{graphicx}%
\usepackage{multirow}%
\usepackage{amsmath,amssymb,amsfonts}%
\usepackage{amsthm}%
\usepackage{mathrsfs}%
\usepackage[title]{appendix}%
\usepackage{xcolor}%
\usepackage{textcomp}%
\usepackage{manyfoot}%
\usepackage{booktabs}%
\usepackage{algorithm}%
\usepackage{algorithmicx}%
\usepackage{algpseudocode}%
\usepackage{listings}%
\usepackage{soul}
\usepackage{tabularx}
\usepackage{bm}
\usepackage{makecell}
\usepackage{xcolor}
\newcommand{\rev}[1]{\textcolor{black}{#1}}

\newcommand{\Qcomp}{\mathbf{Q}_{\mathrm{C}}}
\newcommand{\Qasa}{\mathbf{Q}_{\mathrm{C}}^{\prime}}
\newcommand{\Qupdate}{\widetilde{\mathbf{Q}}_{\mathrm{C}}}

\jyear{2021}%

\theoremstyle{thmstyleone}%
\theoremstyle{thmstyletwo}%

\theoremstyle{thmstylethree}%

\begin{document}

\title[CAM3DNet: Camera-based 3D Object Detection ]{{CAM3DNet: Comprehensively mining the multi-scale features for 3D Object Detection with Multi-View Cameras}}


\author[1,2]{\fnm{Mingxi} \sur{Pang}}

\author*[2]{\fnm{Dingheng} \sur{Wang}}\email{wangdai11@outlook.com}

\author[1]{\fnm{Jian} \sur{Li}}

\author[1]{\fnm{Zhenping} \sur{Sun}}

\author[2]{\fnm{Zekun} \sur{Li}}

\author[2]{\fnm{Zhihang} \sur{Wang}}

\author*[3]{\fnm{Yiqin} \sur{Yang}}\email{yiqin.yang@ia.ac.cn}

\author[3]{\fnm{Bo} \sur{Xu}}

\affil[1]{\orgdiv{College of Intelligent Science and Technology}, \orgname{National University of Defense Technology}, \orgaddress{\city{Changsha} \postcode{410073},  \country{China}}}

\affil[2]{\orgname{Northwest Institute of Mechanical \& Electrical Engineering}, \orgaddress{\city{Xianyang} \postcode{712099},  \country{China}}}

\affil[3]{\orgname{The Key Laboratory of Cognition and Decision Intelligence for Complex Systems, Institute of Automation, Chinese Academy of Sciences}, \orgaddress{\city{Beijing} \postcode{100190},  \country{China}}}


\abstract{

Camera-based 3D object detection is crucial for autonomous driving and mobile robotics due to its semantic richness and cost-effectiveness. However, existing query-based methods still struggle with small objects and targets in cluttered environments. This limitation primarily stems from the insufficient exploitation of dynamic multi-scale information and the inadequate modeling of synergies between object features and query geometry. To address these challenges, we propose CAM3DNet, an efficient sparse query-based framework designed to establish robust global perception through multi-scale image integration. \rev{At its core, CAM3DNet introduces an adaptive self-attention mechanism that incorporates query-dependent, head-specific Gaussian distance priors to model the 3D geometric interactions among composite queries comprising global, temporal, and image-derived information. Furthermore, we extend deformable feature aggregation with a multi-scale hybrid sampling strategy that separately aggregates features from fixed and learnable sampling points before feature-level fusion, thereby improving small-object localization.} Extensive experiments on the nuScenes, Waymo, and Argoverse benchmarks demonstrate the superior effectiveness and generalization of CAM3DNet. 

}




\maketitle

\section{Introduction}\label{sec:introduction}

With the rapid advancement of autonomous driving and mobile robotics, achieving robust 3D perception of the surrounding environment has become increasingly critical for ensuring operational safety and reliability. Compared to LiDAR-based or multi-modal fusion systems, vision-only multi-view setups offer rich semantic representations at a significantly lower deployment cost, closely resembling the primary perceptual paradigm of biological systems \cite{SimLingo,DriveCoT,TGCP}. Consequently, camera-only 3D object detection has attracted substantial research attention, broadly evolving into two dominant streams: convolution-based (conv-based) methods \cite{lss, bevdet, bevdet4d, bevdepth, m2bev, fastbev} and query-based methods \cite{detr3d, deformabledetr, detr, sparse4d}.

The inherent limitations of existing paradigms become particularly evident when dealing with targets of varying scales and heavily cluttered environments. Conv-based methods, while intuitive in generating a dense Bird's-Eye-View (BEV) representation, incur considerable computational redundancy, as large portions of the BEV map cover regions devoid of scene information. More critically, distant or small-scale objects suffer from severe feature degradation during lossy view transformations, making it difficult to balance detection efficiency and sensitivity to small objects. Conversely, query-based methods (e.g., Sparse4D \cite{sparse4d, sparse4dv2,sparse4dv3}) mitigate computational overhead by establishing sparse spatiotemporal correspondences, achieving highly competitive performance. However, these frameworks still struggle with small objects and complex backgrounds. This deficiency primarily stems from their suboptimal exploitation of dynamic multi-scale information and inadequate modeling of the structural relationship between object features and query geometry. For instance, PETR-style approaches \cite{petr,petrv2,StreamPETR,focalpetr} rely on dense global attention but lack fine-grained multi-scale perception; meanwhile, DETR3D-style frameworks \cite{detr, detr3d, deformabledetr,dino,monodetr} offer structural flexibility but often compromise on precision due to sub-optimal 3D anchor initialization and restricted multi-scale feature interactions.

To address these challenges, we propose CAM3DNet, an efficient sparse query-based 3D object detection framework designed to thoroughly mine multi-scale image features and establish robust global perception. \rev{We construct a composite query representation by integrating image-derived adaptive queries, temporally propagated queries, and standard global queries. Rather than introducing new individual query sources, the composite query provides complementary spatial, visual, and temporal information for subsequent query interaction. Building upon these enriched composite representations, we develop an Adaptive Self-Attention (ASA) module as the primary technical contribution of CAM3DNet. ASA introduces query-dependent and head-specific Gaussian distance priors to dynamically regulate query interactions according to their 3D geometric relationships. Furthermore, we extend multi-point deformable feature aggregation through a Multi-Scale Hybrid Sampling (MSHS) strategy, which separately aggregates features from geometry-derived fixed points and query-conditioned learnable points before feature-level fusion.} Extensive experiments on the nuScenes, Waymo Open, and Argoverse 2 benchmarks demonstrate that CAM3DNet effectively balances inference efficiency and detection precision, outperforming most existing state-of-the-art camera-based methods.

We summarize our main contributions as follows:
\begin{itemize}
  \item We propose CAM3DNet, an efficient sparse camera-only 3D object detection framework that  integrates composite queries, adaptive self-attention, and multi-scale hybrid sampling to effectively aggregate dynamic multi-scale object features. \rev{The composite query
  unifies global, temporal, and adaptive queries to provide
  complementary spatiotemporal priors for subsequent query interaction.}
  \item \rev{As the primary technical contribution, we develop an Adaptive Self-Attention (ASA) mechanism that introduces query-dependent and head-specific Gaussian distance priors to dynamically regulate the spatial interaction ranges of heterogeneous queries. To complement ASA, we further extend multi-point deformable feature aggregation through MSHS, which separately aggregates geometric-prior and query-adaptive sampling features before feature-level fusion.}
  \item Extensive evaluations across multiple challenging benchmarks (nuScenes, Waymo, and Argoverse) validate the superior performance and strong generalization capability of CAM3DNet, establishing it as a highly competitive and efficient solution for robust 3D perception.
\end{itemize}


\section{Related Work}\label{sec:related_work}
In this section, we review the literature on multi-view vision-only 3D object detection. Based on their feature representation paradigms, existing methodologies can be broadly categorized into conv-based and query-based approaches.

\subsection{Conv-Based Methods}
Conv-based methods typically focus on transforming multi-view 2D image features into a unified 3D or BEV representation using convolutional operations. Pioneering this field, Lift, Splat, Shoot (LSS) \cite{lss} proposes a probabilistic view transformation technique that predicts categorical depth distributions to project image features into a 3D frustum, which is subsequently splatted onto a dense BEV plane. Building upon this foundation, BEVDet \cite{bevdet} and BEVDet4D \cite{bevdet4d} incorporate multi-frame feature alignment, leveraging spatial correlations from ego-motion to enhance temporal consistency and parallel processing efficiency. To alleviate the reliance on noisy implicit depth estimation, BEVDepth \cite{bevdepth} introduces explicit LiDAR-based depth supervision, significantly boosting projection accuracy. In contrast to these explicit depth-warping methods, alternative approaches such as $M^2$BEV \cite{m2bev} and FastBEV \cite{fastbev} adopt a uniform depth assumption along camera rays. By directly retrieving features from the 3D space using camera intrinsics and extrinsics, these methods circumvent the need for explicit depth estimation, thereby substantially reducing computational overhead while retaining competitive performance.

Although dense BEV representations provide structured spatial context, their computational cost increases with the BEV resolution and perception range. Moreover, the view transformation may weaken fine-grained visual information for small or distant objects.


\subsection{Query-Based Methods}

Query-based methods represent candidate objects using sparse queries and refine them through attention-based feature interaction \cite{attention}. PETR-style methods \cite{petr, petrv2, focalpetr} encode 3D positional information into multi-view image features, while StreamPETR \cite{StreamPETR} propagates object-centric information through a temporal memory queue. DETR3D \cite{detr3d} associates object queries with 3D reference points and projects them onto multi-view features for sampling. BEVFormer \cite{bevformer}, SOLOFusion \cite{solofusion}, and HoP \cite{hop} further aggregate spatial and temporal information through query-based interaction.

The Sparse4D series \cite{sparse4d, sparse4dv2, sparse4dv3} aggregates multi-view, multi-scale, and temporal features around multiple 3D keypoints, while SparseBEV \cite{sparsebev} performs adaptive spatiotemporal sampling without constructing a conventional dense BEV representation. Closely related to our adaptive query formulation, Far3D \cite{far3d} generates 3D queries from 2D proposals and combines them with global queries for long-range detection.

\rev{Monocular depth estimation provides important geometric cues for lifting 2D observations into 3D space. Recent feed-forward foundation models, such as Depth Anything V2 \cite{depthanythingv2} and Metric3D V2 \cite{metric3dv2}, improve cross-domain depth estimation through large-scale training and camera-aware metric modeling. Diffusion-based methods, such as Lotus \cite{lotus} and Iris \cite{iris}, further exploit generative or real-world priors to improve generalization and preserve geometric details. Unlike standalone monocular depth estimation, CAM3DNet uses the estimated depths associated with 2D proposals to initialize adaptive 3D queries.}


\rev{Despite these advances, the joint modeling of heterogeneous queries, 3D geometry, and multi-scale image features remains insufficiently explored. CAM3DNet unifies global, temporal, and image-derived adaptive queries through CQ. Its ASA differs from SparseBEV by applying query-dependent, head-specific Gaussian priors to full 3D distances rather than linear BEV-plane distance modulation. Meanwhile, MSHS fuses fixed- and learnable-point features at the feature level instead of using the hierarchical keypoint aggregation of Sparse4D.}

\section{Methodology}
\label{sec:method}

\begin{figure*}[t]
    \centering
    \includegraphics[width=1.0\linewidth]{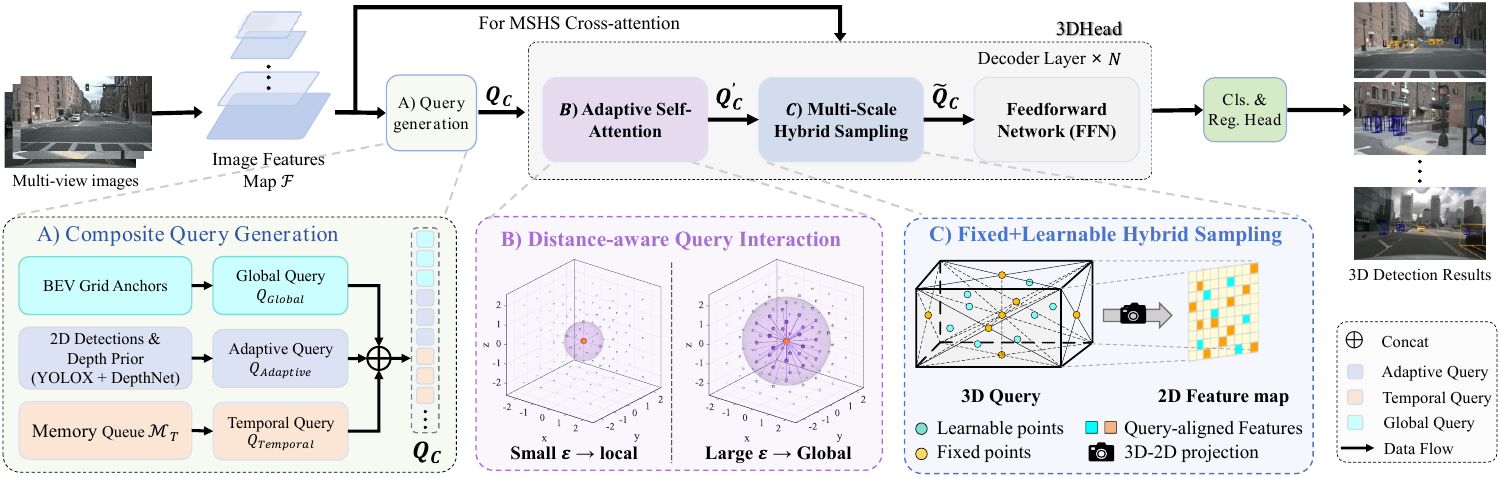}
    \caption{\textbf{The overview of our proposed CAM3DNet}. Feeding the multi-view images into the backbone and FPN neck, we obtain image features $\mathcal{F}$ for query generation and feature interaction. We concatenate the adaptive queries, temporal queries, and 3D global queries as composite queries, and feed them into the Decoder to predict 3D bounding boxes. The decoder repeats $N$ times to produce final predictions.}
    \label{fig:model_framework}
\end{figure*}

\rev{
To address the persistent bottlenecks in multi-scale perception and
query--geometry misalignment, we present CAM3DNet (Fig.~\ref{fig:model_framework}),
an end-to-end framework for robust 3D object detection through dynamic
query interaction and fine-grained feature aggregation. CAM3DNet builds
upon established sparse query-based detectors: the Composite Query (CQ)
unifies standard global queries, StreamPETR-style temporal queries, and
Far3D-style image-derived adaptive queries into a heterogeneous
spatiotemporal representation. Its primary technical contribution is
the Adaptive Self-Attention (ASA) module, which introduces a
query-dependent, head-specific Gaussian distance prior to dynamically
regulate the spatial interaction range among these queries. The
Multi-Scale Hybrid Sampling (MSHS) strategy further builds on the
established deformable-attention operator by combining
geometry-derived fixed edge points with query-conditioned learnable
points for scale-aware image-feature aggregation. The complete
execution pipeline is summarized in Algorithm~\ref{alg:cam3dnet}.
}

\begin{algorithm}
\caption{CAM3DNet}\label{alg:cam3dnet}
\begin{algorithmic}[1]
\Require Multi-view images $\mathcal{I}$, camera parameters $K$ and \rev{$T$}, number of decoder layers $N$
\Ensure 3D bounding boxes $\mathcal{B}$ and class scores $\mathcal{S}$
\State Extract multi-scale image features $\mathcal{F} = \{ \mathcal{F}^{1/4}, \mathcal{F}^{1/8}, \mathcal{F}^{1/16}, \mathcal{F}^{1/32} \}$ from $\mathcal{I}$;
\State Initialize global queries $Q_\mathrm{Global}$;
\State Generate adaptive queries $Q_\mathrm{Adaptive}$ from 2D detection with Eq.~\ref{eq:3};
\State Propagate temporal queries $Q_\mathrm{Temporal}$ from memory queue $\mathcal{M}_\mathrm{T}$ with Eq.~\ref{eq:4};
\State Construct composite queries 
$Q_\mathrm{C}=\text{Concat}(Q_\mathrm{Adaptive}, Q_\mathrm{Temporal}, Q_\mathrm{Global})$ (Eq.\ref{eq:1} in Section~\ref{ssec:query});

\For{$n = 1$ to $N$}
    \State Leveraging query interactions within $Q_\mathrm{C}$ to update $Q_\mathrm{C}^\prime = \mathrm{ASA}(Q_\mathrm{C}; \mathbf{D})$
    (Eqs.~\ref{eq:7}--\ref{eq:7-1} in Section~\ref{ssec:asa});

    \State Project the fixed and learned 3D sampling points onto the corresponding image planes to obtain \rev{$P^{2D} = \mathbf{K} \cdot \mathbf{T} \cdot \left( P_\mathrm{fixed} , P_\mathrm{learned} \right)$}(Eqs.~\ref{eq:9}--\ref{eq:10} in Section~\ref{ssec:mshs});
    
    \State Sample image features $\mathcal{F}$ and interact with $Q_\mathrm{C}^\prime$ to update 
    \rev{$\widetilde Q_{\mathrm C}=\alpha\,\mathcal D  \bigl(P_{\mathrm{fixed}}^{\mathrm{2D}}\bigr)+(1-\alpha)\,\mathcal D \bigl(P_{\mathrm{learned}}^{\mathrm{2D}}\bigr).$} (Eq.~\ref{eq:11} in Section~\ref{ssec:mshs});
    
    \State Refine queries $\tilde{Q}_\mathrm{C}$ with FFN;

\EndFor

\State Update memory queue $\mathcal{M}_\mathrm{T}$;
\State Decode final 3D bounding boxes $\mathcal{B}$ and class scores $\mathcal{S}$;
\State \Return $\mathcal{B}, \mathcal{S}$

\end{algorithmic}
\end{algorithm}

\subsection{Composite Query Generation}
\label{ssec:query}
In multi-view 3D object detection, localizing small objects remains a formidable challenge. These targets occupy limited image regions and suffer from severe feature degradation as the observation distance increases. To tackle this issue, we formulate a composite query comprising three complementary components: an adaptive query $Q_\mathrm{Adaptive}$, a temporal query $Q_\mathrm{Temporal}$, and a global query $Q_\mathrm{Global}$. Each query instance is parameterized by a 9-dimensional vector in the BEV space, encompassing its 3D geometric center $(x,y,z)$, geometric size $(w,l,h)$, orientation angle $\theta$, and velocity $(v_\mathrm{x},v_\mathrm{y})$. The final composite query, $Q_\mathrm{C}$, is constructed by concatenating these three components:
\begin{equation}
\label{eq:1}
    Q_\mathrm{C}=\text{Concat}(Q_\mathrm{Adaptive}, Q_\mathrm{Temporal}, Q_\mathrm{Global}).
\end{equation}

\rev{Throughout this paper, $\Qcomp$, $\Qasa$, and $\Qupdate$ denote the initial composite queries, the queries updated by adaptive self-attention, and the queries updated through image-feature interaction, respectively.}

For the global query, following standard practices in BEV-based 3D object detection \cite{StreamPETR,sparse4dv2,sparsebev}, we randomly initialize $n$ 9-dimensional vectors within a predefined 3D grid space. Subsequently, we project them into high-dimensional representations via sinusoidal positional encoding \cite{attention}, yielding the global queries $Q_\mathrm{Global}$, as depicted in Fig.~\ref{fig: 31}(a).

\subsubsection{Adaptive Query Generation}
\label{subsub aq}
\begin{figure}[t]
\centering\includegraphics[width=1.0\linewidth]{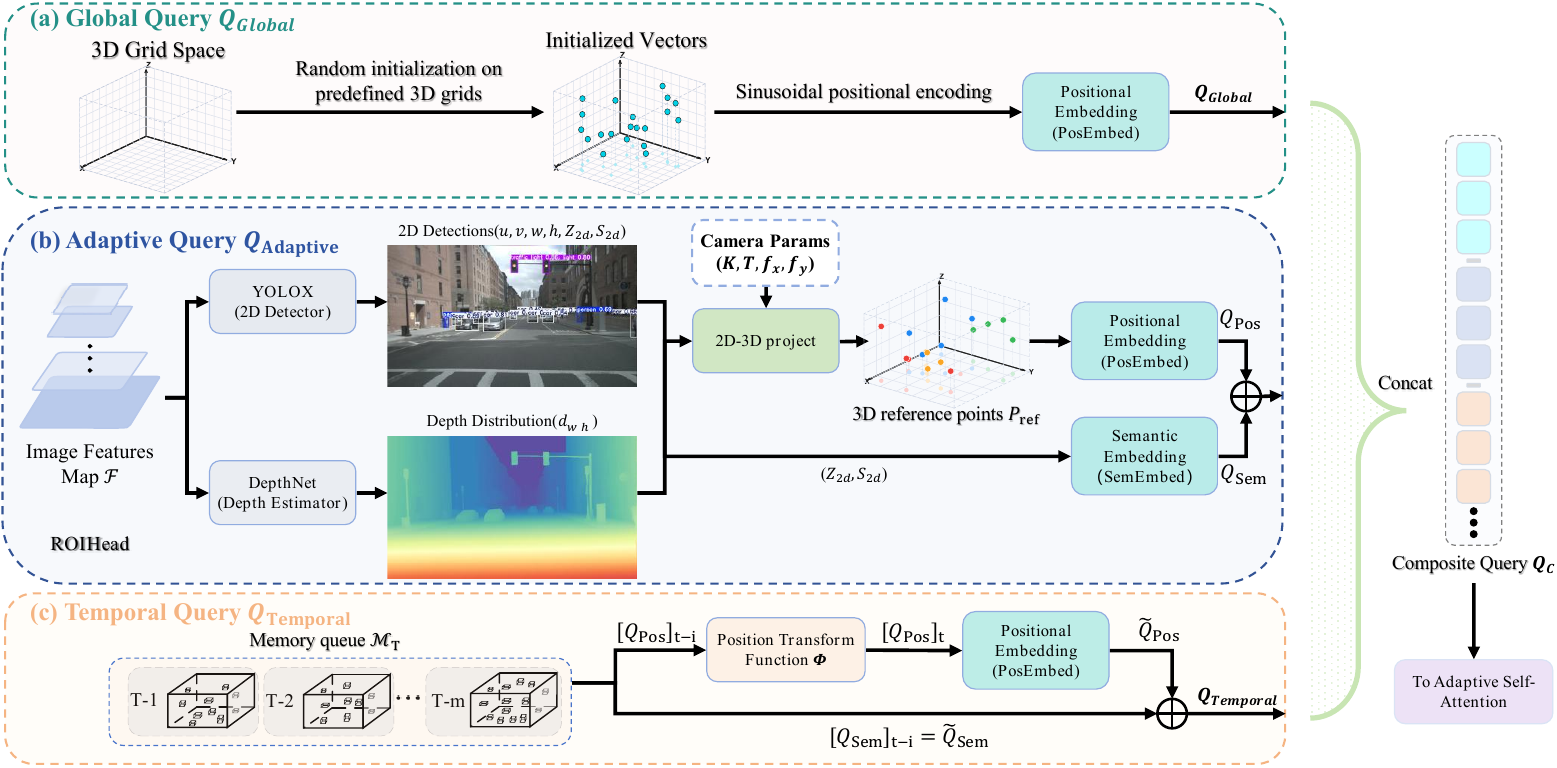}
\caption{\textbf{Composite query generation framework}. The proposed composite query consists of three complementary components: (a) global query $Q_\mathrm{Global}$, (b) adaptive query $Q_\mathrm{Adaptive}$ with ROIHead, (c) temporal query $Q_\mathrm{Temporal}$.}
\label{fig: 31}
\end{figure}

As illustrated in Fig.~\ref{fig: 31}(b), the adaptive query $Q_\mathrm{Adaptive}$ combines 2D detection results with explicit depth estimation. Given the multi-scale features $\mathcal{F}$ at 1/4, 1/8, 1/16, and 1/32 resolutions, the YOLOX-based ROIHead \cite{yolox} predicts 2D bounding boxes $(u,v,w,h)$ and their confidence scores. Meanwhile, DepthNet, inspired by Mono-DETR \cite{monodetr}, predicts a categorical depth distribution and decodes the corresponding depth $d_\mathrm{wh}$.

\rev{For all datasets, the 2D targets are generated by projecting the original 3D ground-truth boxes onto the corresponding camera views, with class labels inherited from the 3D annotations. Therefore, the ROIHead introduces an auxiliary 2D detection loss but requires no additional manually annotated 2D labels. It is jointly trained with CAM3DNet from random initialization, without using external 2D data or pretrained detector weights.
}

\rev{To dynamically determine the query number, we first apply $3\times3$ local-maximum suppression to the confidence maps predicted by the ROIHead. Let $C_{\max}$ denote the maximum number of local maxima with confidence scores above $\tau=0.1$ among all camera views in the current batch. The shared Top-$k$ value and the resulting number of adaptive queries are determined as
\begin{equation}
\label{eq:adaptive_query_number}
k=\min(C_{\max},k_{\max}), 
\qquad n=Mk,
\end{equation}
where $k_{\max}=100$ and $M$ is the number of cameras. Each camera independently selects its Top-$k$ candidates using the same $k$. Therefore, the query number dynamically adapts to the current predictions during both training and inference. The determination of $k$ does not require ground-truth annotations.}

The selected 2D candidates are then associated with their estimated depths and projected into the 3D space. For each candidate, the 3D reference point is formulated as:
\begin{equation}
\label{eq:2}
{P_\mathrm{ref}}=\left[
\begin{aligned}
 & {P_\mathrm{xyz}}=T^{-1}K^{-1}{{\left[ u \cdot {d_\mathrm{wh}}, v \cdot {d_\mathrm{wh}}, {d_\mathrm{wh}}, 1 \right]}^{\rm T}}, \\ 
 & {P_\mathrm{wh}}={d_\mathrm{wh}}{{[{w},{h}]}^{\rm T}}/{{[{f_\mathrm{x}},{f_\mathrm{y}}]}^{\rm T}}, \\
 & {P_\mathrm{l}}={P_\mathrm{w}}, \; {P_\mathrm{\theta }}=0, \; {P_\mathrm{{{v}_{x}}{{v}_{y}}}}=0 \\ 
\end{aligned}
\right] \in \mathbb{R}^{9}
\end{equation}
\rev{where $K$ and $T$ denote the intrinsic matrix and extrinsic transformation of the corresponding camera, respectively.} Furthermore, ${P_\mathrm{xyz}, P_\mathrm{whl}, P_\mathrm{v_{x}v_{y}}, P_\mathrm{\theta}}$ denote the projected 3D geometric center, 3D geometric size, velocity along the $x$ and $y$ axes, and orientation angle in the BEV space, respectively. 

After explicitly projecting the 2D bounding boxes into the 3D space to establish ${P_\mathrm{ref}}$, we obtain the final adaptive query $Q_\mathrm{Adaptive}$ via:
\begin{equation}
\label{eq:3}
\begin{aligned}
  & ({Z}_\mathrm{2d},{S}_\mathrm{2d}) = \text{YOLOX}(\mathcal{F}), \\
  & {{Q}_\mathrm{Pos}}=\text{PosEmbed}({P_\mathrm{ref}}), \\ 
  & {{Q}_\mathrm{Sem}}=\text{SemEmbed}({Z}_\mathrm{2d},{S}_\mathrm{2d}), \\ 
  & {{Q}_\mathrm{Adaptive}}={{Q}_\mathrm{Pos}}+{{Q}_\mathrm{Sem}},
\end{aligned}
\end{equation}
where $\text{PosEmbed}(\cdot)$ is the positional encoding function utilizing sinusoidal embeddings \cite{cosine_anneal} to generate positional features $Q_\mathrm{Pos}$. The variables $(Z_\mathrm{2d}, S_\mathrm{2d})$ represent the semantic features and confidence scores derived from the 2D detection branch, respectively. Additionally, $\text{SemEmbed}(\cdot)$ serves as the semantic encoding function, employing a Multi-Layer Perceptron (MLP) to extract refined semantic embeddings $Q_\mathrm{Sem}$.

Fundamentally, the entire generation process of $Q_\mathrm{Adaptive}$ is fully differentiable, allowing the network to continuously refine query precision during training. Because the formulation explicitly incorporates multi-scale 2D features and rigorous 2D-to-3D geometric projections, we argue that $Q_\mathrm{Adaptive}$ encapsulates rich, scale-aware object priors, effectively enhancing the model's sensitivity to geometrically challenging targets.

\subsubsection{Temporal Query Generation}
\label{subsub tq}

\rev{Inspired by recent BEV perception methods \cite{sparse4dv2, StreamPETR}, we propagate historical query states to construct temporal queries. Each query is decomposed into a geometric component ${\widetilde{Q}_\mathrm{Pos}}$ and a semantic component ${\widetilde{Q}_\mathrm{Sem}}$. At the initial timestamp, these components are initialized from $\widetilde{Q}_{\mathrm{Adaptive}}$. For subsequent frames, the semantic component is preserved, while the geometric component is updated by a positional prediction model using the time interval and ego-motion transformation.}


\rev{We maintain $\mathcal{M}_{\mathrm{T}}$ as a strict first-in-first-out (FIFO) queue with a capacity of 1,024 query states, corresponding to four groups of 256 queries, as illustrated in Fig.~\ref{fig: 31}(c). Before processing the current frame, the oldest group of 256 queries is dequeued and propagated to the current timestamp, while the other entries remain in their original order for subsequent propagation. After decoding the current frame, the 256 valid decoder queries with the highest classification confidence are appended to the rear of the queue. The same FIFO update strategy is used during training and inference.}

The features ${\widetilde{Q}}_\mathrm{Pos}$ and ${\widetilde{Q}}_\mathrm{Sem}$ propagated from frame $t-i$ to the current frame $t$ are formulated as follows:

\begin{equation}
\label{eq:4}
\begin{aligned}
{\widetilde{Q}}_\mathrm{Pos} &=  \Phi(\left[{Q_\mathrm{Pos}}\right]_\mathrm{t-i})= \left[
\begin{aligned}
  & {{\left[ x,y,z \right]}_\mathrm{t}}={{\mathbf{R}}_\mathrm{t-i\to t}}{{\left( \left[ x,y,z \right]+\Delta t\left[ {{v}_\mathrm{x}},{{v}_\mathrm{y}},{{v}_\mathrm{z}} \right] \right)}_\mathrm{t-i}}+{{\mathbf{T}}_\mathrm{t-i\to t}}, \\ 
 & {{\left[ w,l,h \right]}_\mathrm{t}}={{\left[ w,l,h \right]}_\mathrm{t-i}}, \\ 
 & {{\left[ {{v}_\mathrm{x}},{{v}_\mathrm{y}} \right]}_\mathrm{t}}={{\left[ {{v}_{x}},{{v}_{y}} \right]}_\mathrm{t-i}}, \\ 
 & {{\left[ \theta \right]}_\mathrm{t}}={{\mathbf{R}}_\mathrm{t-i\to t}}{{\left[ \theta \right]}_\mathrm{t-i}} \\ 
\end{aligned}
\right], \\
{{\widetilde{Q}}_\mathrm{Sem}} &= \left[{Q_\mathrm{Sem}}\right]_\mathrm{t-i},
\end{aligned}
\end{equation}
where $i\in \{1,\cdots ,L\}$ denotes the frame index within the memory queue. The terms ${\mathbf{R}}_\mathrm{t-i\to t}$ and ${\mathbf{T}}_\mathrm{t-i\to t}$ represent the rotation and translation matrices derived from the vehicle's ego-motion between frame $t-i$ and frame $t$, respectively. Other geometric notations are defined in Section~\ref{ssec:query}.

Similar to the construction of $Q_\mathrm{Adaptive}$, we fuse the updated geometric and semantic features to obtain the final temporal query ${Q}_\mathrm{Temporal}$:
\begin{equation}
    {{Q}_\mathrm{Temporal}}={{\widetilde{Q}}_\mathrm{Pos}}+{{\widetilde{Q}}_\mathrm{Sem}}.
\end{equation}
Through this mechanism, we effectively extend the multi-scale representation capabilities into the temporal domain. Because the queries at every timestamp originate from $Q_\mathrm{Adaptive}$, they consistently inherit rich, scale-aware 2D visual priors from the multi-view cameras.

\subsection{Adaptive Self-Attention}
\label{ssec:asa}
\begin{figure}[t]
\centering\includegraphics[width=0.5\linewidth]{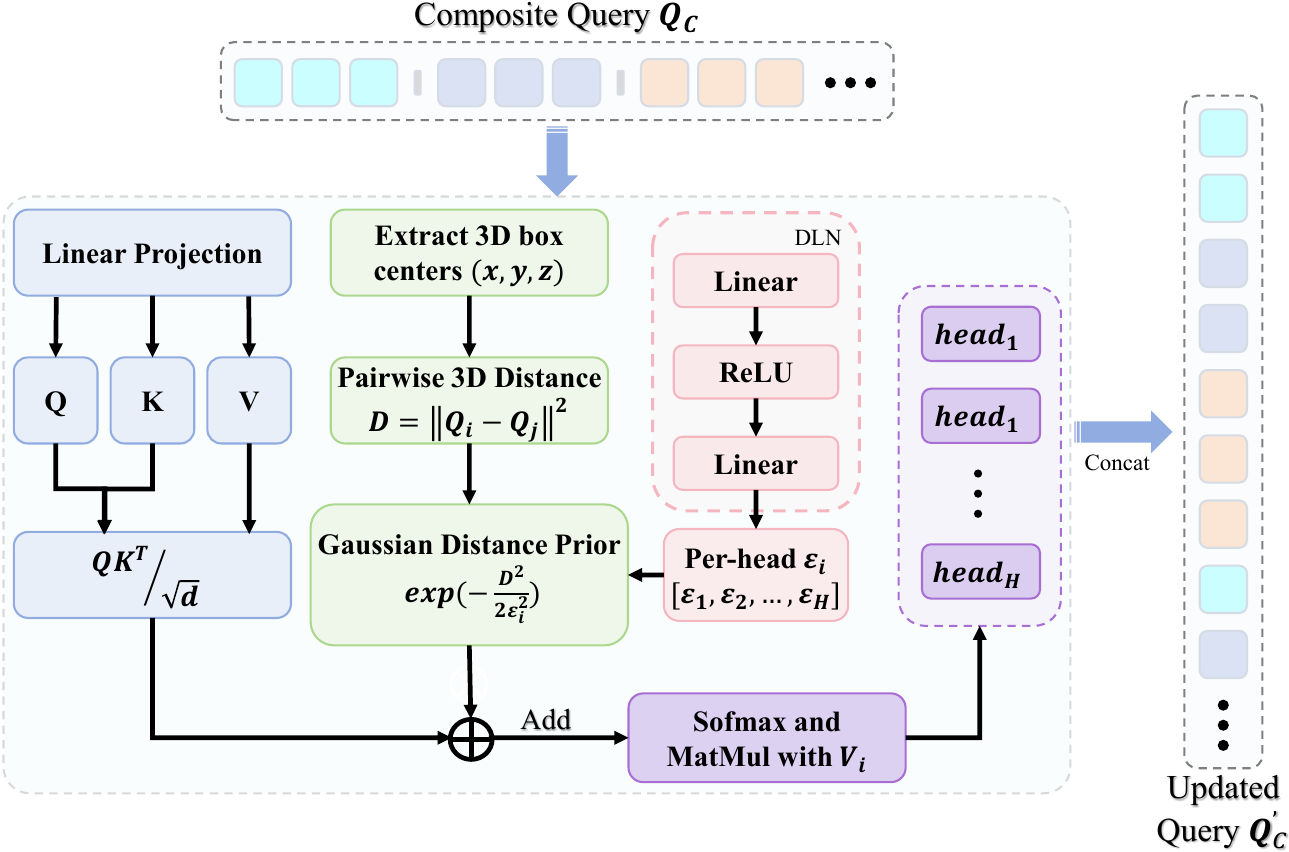}
\caption{\textbf{Adaptive Self-attention framework}. We introduce a scale-adaptive weighting ${\varepsilon}$ by employing a two-layer linear network. This parameter is used to incorporate distance information into the attention weights, enabling an adaptive receptive field.}
\label{fig:asa_fig}
\end{figure}

Leveraging the highly enriched representations encapsulated within the composite query $Q_\mathrm{C}$, the subsequent critical step is to enhance the model's capability to effectively aggregate structural context. Inspired by 2D detection priors, we observe that the semantic correlation between distinct image regions naturally attenuates as their spatial distance increases. To fully exploit the expressive capacity of composite queries while filtering out noisy long-range interactions, we introduce an Adaptive Self-Attention (ASA) mechanism, as illustrated in Fig.~\ref{fig:asa_fig}. 
 
Initially, the ASA module computes the pairwise 3D spatial distances between all queries to formulate a structural distance matrix. Unlike existing methods that rely on pillar-based prediction references, we explicitly adopt 3D anchor coordinates. Specifically, the Euclidean distance $D_\mathrm{i,j}$ between the 3D coordinates $(x_\mathrm{i}, y_\mathrm{i}, z_\mathrm{i})$ and $(x_\mathrm{j}, y_\mathrm{j}, z_\mathrm{j})$ of the $i$-th and $j$-th queries is defined as ${D_\mathrm{i,j}}=\sqrt{{(x_\mathrm{i}-x_\mathrm{j})}^{2} + {(y_\mathrm{i}-y_\mathrm{j})}^{2} + {(z_\mathrm{i}-z_\mathrm{j})}^{2}}$.

Subsequently, we integrate this structural distance prior $\mathbf{D}$ into the self-attention mechanism via a Gaussian penalty kernel to explicitly model scale-aware spatial relationships. Formally, the output of the $i$-th attention head, denoted as $\text{head}_\mathrm{i}$, is mathematically formulated as:
\rev{
\begin{equation}
\label{eq:7}
\begin{aligned}
\text{head}_\mathrm{i}
&= \text{ASA}(Q_\mathrm{C,i}, K_\mathrm{i}, V_\mathrm{i}) \\
&= \operatorname{Softmax}\left(
\frac{Q_\mathrm{C,i} K_\mathrm{i}^\top}{\sqrt{d}}
+
\exp\left(-\frac{\mathbf{D}^2}{2\varepsilon_\mathrm{i}^2}\right)
\right)V_\mathrm{i},
\end{aligned}
\end{equation}
}

\rev{where $i\in\{1,\ldots,\mathrm{H}\}$ indexes the attention heads and $d$ denotes the channel dimension per head. $\mathbf{D}$ is the pairwise Euclidean distance matrix between the 3D reference points. The second additive term inside Softmax acts as a head-specific Gaussian spatial bias, whose spatial extent is controlled by the positive learnable bandwidth $\varepsilon_i$. A smaller $\varepsilon_i$ favors more localized interactions, whereas $\varepsilon_i\rightarrow\infty$ makes the bias uniform and recovers vanilla self-attention.}

The updated composite query $Q_{\mathrm{C}}^\prime$ is then obtained by concatenating the outputs from all $\mathrm{H}$ heads:
\begin{equation}
\label{eq:7-1}
Q_{\mathrm{C}}^\prime = \text{Concat}\left(\text{head}_\mathrm{1}, \text{head}_\mathrm{2}, \cdots, \text{head}_\mathrm{H}\right).
\end{equation}

Crucially, the query-specific bandwidth parameters $[{\varepsilon}_\mathrm{1}, {\varepsilon}_\mathrm{2}, \cdots , {\varepsilon}_\mathrm{H}]$ in Eq.~\ref{eq:7} are dynamically predicted from the input query $Q_\mathrm{C}$ as follows:
\begin{equation}
\label{eq:8}
\begin{aligned}
    [{\varepsilon}_\mathrm{1}, {\varepsilon}_\mathrm{2}, \cdots , {\varepsilon}_\mathrm{H}] &=
    \begin{cases} 
    \text{Linear}(Q_\mathrm{C})_{d\to \mathrm{H}} \\
    \quad \text{or} \\
    \text{MLP}(Q_\mathrm{C})=\text{Linear}\left( \text{ReLU}\left( \text{Linear}(Q_\mathrm{C}) \right) \right)_{d\to \mathrm{H}}
    \end{cases} 
\end{aligned}
\end{equation}
where the projection weights are shared across all queries during the training process. Here, $\text{Linear}(\cdot)$ denotes a single linear projection layer, and $\text{MLP}(\cdot)$ represents a two-layer Multi-Layer Perceptron.


\begin{figure}[t]
\centering
\includegraphics[width=1.0\linewidth]{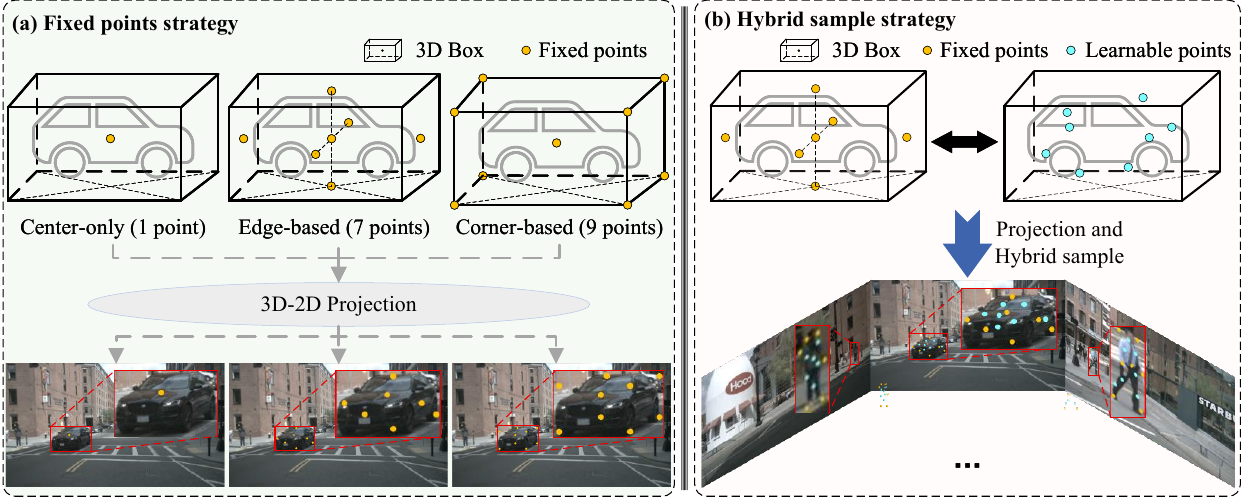}
\caption{\textbf{The fixed and learnable points for sampling}. (a) Illustration of the three different fixed-point setting strategies. (b) Illustration of the proposed MSHS strategy, which combines fixed and learnable points to generate hybrid sampling points that are further projected into the 2D image plane for feature sampling. }
\label{fig:sample_points}
\end{figure}

\subsection{Multi-Scale Hybrid Sampling}
\label{ssec:mshs}
To improve the detection of small objects within cluttered scenes, we introduce a Multi-Scale Hybrid Sampling (MSHS) strategy designed to extract comprehensive, object-level representations. Following the adaptive self-attention phase, we employ a sparse cross-attention module—grounded in the deformable attention mechanism \cite{deformabledetr}—to iteratively fuse the queries with multi-scale image features. Relying on a single 3D reference point for each query in $Q_\mathrm{C}^\prime$ is inherently restrictive for capturing rich geometric context. Therefore, we aim to expand the spatial footprint during the point sampling process. As illustrated in Fig.~\ref{fig:sample_points}(b), we formulate the 3D sampling point set $P^\mathrm{3D} \in \mathbb{R}^{N_\mathrm{q} \times 3}$ (where $N_\mathrm{q}$ denotes the number of queries) as a balanced combination of fixed geometric points $P_\mathrm{fixed}$ and dynamically learnable points $P_\mathrm{learned}$. The hybrid sampling computation is defined as follows:
\begin{equation}
\label{eq:9}
\begin{aligned}
    &P_\mathrm{learned} = \text{Linear}(Q_\mathrm{C}^\prime), \\
    &\rev{P^\mathrm{3D} = \left(P_\mathrm{fixed} , P_\mathrm{learned}\right),}
\end{aligned}
\end{equation}
where $P_\mathrm{fixed}$ are deterministically located at the keypoints of a 3D bounding box prior, as shown in Fig.~\ref{fig:sample_points}(a). Conversely, the learned sampling points $P_\mathrm{learned}$ are generated from the query $Q_\mathrm{C}^\prime$ via a linear projection layer. $\mathcal{P}^{\mathrm{3D}}$ denotes the hybrid sampling-point collection composed of two branch-specific tensors.

Next, the hybrid reference points $P^\mathrm{3D}$ are projected onto the 2D image plane to obtain the 2D sampling points $P^\mathrm{2D}$. We utilize the camera \rev{intrinsic matrix $K$ and the extrinsic matrix $T$} to perform this projection:
\begin{equation}
\label{eq:10}
    \rev{P^\mathrm{2D} = K \cdot T \cdot P^\mathrm{3D}.}
\end{equation}

\rev{Finally, given the separately projected fixed and learnable sampling points, $P_\mathrm{fixed}^{\mathrm{2D}}$ and $P_\mathrm{learned}^{\mathrm{2D}}$, together with the multi-scale image features $\mathcal{F}$, we employ two independently parameterized multi-head deformable-attention branches to aggregate geometric-prior and query-adaptive visual features:}

\rev{
\begin{equation}
\label{eq:11}
\begin{aligned}
\Delta P_{hqs}
    &= \operatorname{Linear}_{\mathrm{offset}}
       \bigl(Q_{\mathrm C}'\bigr),
\\
A_{hqs}
    &= \operatorname{Softmax}_{s}
       \left(
       \operatorname{Linear}_{\mathrm{attn}}
       \bigl(Q_{\mathrm C}'\bigr)
       \right),
\\
\mathcal D\bigl(P^{\mathrm{2D}}\bigr) &= \text{DeformAttn}\left( Q_\mathrm{C}^\prime, P^\mathrm{2D}, \mathcal{F} \right) \\
    &=
    \sum_{h=1}^{H}
    \mathbf W_h
    \left[
    \sum_{s=1}^{S}
    A_{hqs}\,
    \mathbf W_h'
    \mathcal F
    \bigl(P^{\mathrm{2D}}+\Delta P_{hqs}\bigr)
    \right],
\\
\widetilde Q_{\mathrm C}
    &=
    \alpha\,\mathcal D
    \bigl(P_{\mathrm{fixed}}^{\mathrm{2D}}\bigr)
    +(1-\alpha)\,\mathcal D
    \bigl(P_{\mathrm{learned}}^{\mathrm{2D}}\bigr).
\end{aligned}
\end{equation}
}

\rev{where $\alpha$ is a learnable gating parameter for feature fusion, $h$ and $s$ index the attention head and sampled point, respectively}, $H$ and $S$ denote the total numbers of attention heads and sampled points per head, respectively, $\mathbf{W}_{h}$ and $\mathbf{W}_{h}^{\prime}$ are the learnable output and value projection matrices, and the sampling offsets $\Delta P_{hqs}$ and attention weights $A_{hqs}$ are dynamically predicted from $Q_{\mathrm C}^{\prime}$ through the corresponding linear transformations. \rev{The updated query $\widetilde Q_{\mathrm C}$ is obtained by fusing independently aggregated geometric-prior and query-adaptive features through $\alpha$, without interpolating coordinates or requiring point-wise correspondence between the two point sets.}


Subsequently, the aggregated query $\widetilde{Q}_\mathrm{C}$ is refined by a two-layer FFN with a hidden dimension of 2048. As shown in Fig.~\ref{fig:model_framework}, the decoder block is repeated $N$ times, after which the final output is fed into the classification and regression heads to produce the 3D bounding-box predictions.

\subsection{Implementation Details}
\label{ssec:implementation}
The overall architecture of the proposed CAM3DNet comprises four primary components: a backbone network, a Feature Pyramid Network (FPN), an ROIHead, and a 3D Detection Head(3DHead). To extract robust visual representations, we employ ResNet-50, ResNet-101 \cite{resnet}, or VoVNetV2-99 (V2-99) \cite{v2-99} as the backbone, depending on the specific experimental configuration. Subsequently, the FPN processes these extracted features to generate a hierarchical feature pyramid consisting of four scales, with spatial resolutions downsampled by factors of $4\times, 8\times, 16\times,$ and $32\times$ relative to the original input image.


\rev{The model is trained for 24 epochs on three NVIDIA RTX 3090 GPUs using AdamW~\cite{adamw}, with a global batch size of 6 and an initial learning rate of $4\times10^{-4}$. CBGS~\cite{cbgs} is disabled in all experiments to isolate the effectiveness of our method. We use 644 global queries, 256 temporal queries, and $n$ adaptive queries, resulting in $900+n$ composite queries per forward pass. To assess sensitivity to random seeds, we train the ResNet-50 model independently with seeds 0, 1, and 2 under identical settings. Each run is evaluated using its final-epoch checkpoint, and the results are reported as the mean and sample standard deviation.}

\section{Experiments}


In this section, we conduct comprehensive experiments to rigorously validate the effectiveness of the proposed CAM3DNet. 
Specifically, we benchmark our approach on three large-scale, widely adopted 3D object detection datasets: nuScenes \cite{nuscenes}, Waymo Open Dataset \cite{waymo}, and Argoverse 2 \cite{argoverse}. 

\begin{table*}[t]
\centering
\caption{Comparison on the nuScenes validation split. $^{\ast}$ benefited from the perspective-view pre-training. $^{\dagger}$ uses 900 anchors initialized from k-means clustering of the nuScenes train set. \rev{The ResNet-50 result of CAM3DNet in this table is obtained using the default seed 0; repeated-run statistics are reported in Table~\ref{tab:seed_results}.} \label{tab:nusc_val}}
\resizebox{\textwidth}{!}{
\begin{tabular}{c|cc|cc|ccccc|cc}
\hline
Method & Backbone & Image Size & 
mAP$\uparrow$ & NDS$\uparrow$ &
mATE$\downarrow$ & mASE$\downarrow$ & mAOE$\downarrow$ & mAVE$\downarrow$ & mAAE$\downarrow$ & FLOPs(G) & Params(M) \\

\hline

$\text{Focal-PETR\cite{focalpetr} }$ & ResNet50-DCN &$320\times 800$ &
0.320 & 0.381 & 0.788 & 0.278 & 0.595 & 0.893 & 0.228 & - & -\\

$\text{PETRv2\cite{petrv2} }$ & ResNet50 &$256\times 704$ &
0.349 & 0.456 & 0.700 & 0.275 & 0.580 & 0.437 & 0.187 & - & - \\

$\text{SOLOFusion\cite{solofusion} }$ & ResNet50 &$256\times 704$ & 0.427 & 0.534 & \textbf{0.567} & 0.274 & 0.511 & 0.252 & 0.181 & - & - \\

$\text{StreamPETR\cite{StreamPETR}}^{\ast}$ & ResNet50 & $256\times 704$ &
0.450 & 0.550 & 0.613 & {0.267} & 0.413 & 0.265 & 0.196 & \textbf{110.046} & \textbf{34.44} \\

$\text{Sparse4D v2\cite{sparse4dv2}}^{\dagger}$ & ResNet50 & $256\times 704$ &
0.439 & 0.539 & 0.598 & 0.270 & 0.475 & 0.282 & \textbf{0.179} & - & - \\

$\text{SparseBEV\cite{sparsebev}}^{\ast}$ & ResNet50 & $256\times 704$ & 0.448 & \textbf{0.558} & 0.581 & 0.271 & {0.373} & \textbf{0.247} & 0.190 & 244.313 & 44.07 \\[0.25ex]

$\text{Far3D\cite{far3d}}^{\ast}$ & ResNet50 & $256\times 704$ & 0.4441 & 0.5407 & 0.604 & 0.269 & 0.466 & 0.280 & 0.1954 & 203.825 & 43.004 \\[0.25ex]

$\text{MDHA-conv\cite{mdha}}$ & ResNet50 & $256\times 704$ & 0.396 & 0.498 & 0.681 & 0.276 & 0.517 & 0.338 & 0.183 & - & - \\[0.25ex]

$\text{GeoBEV\cite{geobev}}$ & ResNet50 & $256\times 704$ & 0.415 & 0.535 & 0.533 & 0.265 & 0.419 & 0.298 & 0.214 & - & - \\[0.25ex]

$\text{DenseBEV\cite{densebev}}$ & ResNet50 & $256\times 704$ & 0.449 & 0.549 & 0.615 & \textbf{0.264} & \textbf{0.330} & 0.360 & 0.189 & - & - \\[0.25ex]

$\text{CAM3DNet(Ours)}^{\ast}$ & ResNet50 & $256\times 704$ & \textbf{0.4598} & 0.5511 & 0.5994 & 0.269 & 0.4427 & 0.2792 & 0.1978 & 194.217 & 42.997 \\

\hline


\rule{0pt}{2.5ex}$\text{StreamPETR\cite{StreamPETR}}$ & V2-99 & $320\times 800$ & 0.482 & 0.571 & 0.610 & \textbf{0.256} & 0.375 & 0.263 & 0.194 & \textbf{740.67426} & \textbf{80.517} \\[0.25ex]

$\text{Stream3DPPE\cite{3dppe}}$ & V2-99 & $320\times 800$ & 0.500 & 0.585 & 0.565 & 0.261 & 0.376 & \textbf{0.251} & 0.200 & - & - \\[0.25ex]

$\text{CAM3DNet(Ours)}^{\ast}$ & V2-99 & $320\times 800$ & \textbf{0.5168} & \textbf{0.6021} & \textbf{0.5601} & 0.2568 & \textbf{0.2938} & 0.2594 & \textbf{0.1932} & 824.88226 & 89.066\\

\hline

$\text{Focal-PETR\cite{focalpetr} }$ & ResNet101-DCN &$512\times 1408$ &
0.390 & 0.461 & 0.678 & 0.263 & 0.395 & 0.804 & 0.202 & - & - \\

$\text{PETRv2\cite{petrv2} }$ & ResNet101 &$640\times 1600$ &
0.421 & 0.524 & 0.681 & 0.267 & 0.357 & 0.377 & 0.186 & - & - \\

$\text{SOLOFusion\cite{solofusion} }$ & ResNet101 &$512\times 1408$ & 0.483 & 0.582 & {0.503} & 0.264 & 0.381 & 0.246 & 0.207 & - & - \\

$\text{StreamPETR\cite{StreamPETR}}^{\ast}$ & ResNet101 & $512\times 1408$ &
0.504 & 0.592 & 0.569 & 0.262 & \textbf{0.315} & 0.257 & 0.199 & \textbf{697.733} & \textbf{61.937} \\

$\text{Sparse4D v2\cite{sparse4dv2}}^{\dagger}$ & ResNet101 & $512\times 1408$ &
0.505 & 0.594 & 0.548 & 0.268 & 0.348 & 0.239 & 0.184 & - & - \\

$\text{SparseBEV\cite{sparsebev}}^{\ast}$ & ResNet101 & $512\times 1408$ & 0.501 & 0.592 & 0.562 & 0.265 & 0.321 & 0.243 & 0.195 & 832.002 & 63.182 \\[0.25ex]

$\text{Far3D\cite{far3d}}^{\ast}$ & ResNet101 & $512\times 1408$ & 0.510 & 0.594 & 0.551 & {0.258} & 0.372 & \textbf{0.238} & 0.195 & 791.512 & 61.945 \\[0.25ex]

$\text{MDHA-conv\cite{mdha}}$ & ResNet101 & $512\times 1408$ & 0.464 & 0.550 & 0.608 & 0.261 & 0.444 & 0.321 & 0.184 & - & - \\[0.25ex]

$\text{TiGDistill-BEV\cite{TiGDistill-BEV}}$ & ResNet101 & $512\times 1408$ & 0.479 & 0.582 & \textbf{0.498} & \textbf{0.254} & 0.335 & 0.285 & 0.204 & - & - \\[0.25ex]

$\text{GeoBEV\cite{geobev}}$ & ResNet101 & $512\times 1408$ & 0.464 & 0.550 & 0.608 & 0.261 & 0.444 & 0.321 & 0.184 & - & - \\[0.25ex]

$\text{HV-BEV\cite{hvbev}}$ & ResNet101 & $900\times 1600$ & 0.439 & 0.533 & 0.617 & 0.264 & 0.388 & 0.375 & \textbf{0.127} & - & - \\[0.25ex]

$\text{CAM3DNet(Ours)}^{\ast}$ & ResNet101 & $512\times 1408$ & \textbf{0.5124} & \textbf{0.600} & 0.5482 & 0.2631 & 0.3260 & 0.2411 & {0.1826} & 781.905 & 61.938 \\

\hline

\end{tabular}
}
\end{table*}

\begin{table*}[t]
\centering
\caption{ Comparison on the nuScenes test split. 'C' and 'R' represent camera and radar, respectively. $^{\dagger}$ uses 900 anchors initialized from k-means clustering of the nuScenes train set.\label{tab:nusc_test}}
\resizebox{\textwidth}{!}{
\begin{tabular}{c|cc|c|cc|ccccc}

\hline
Method & Backbone & Image Size & Input &
mAP$\uparrow$ & NDS$\uparrow$ &
mATE$\downarrow$ & mASE$\downarrow$ & mAOE$\downarrow$ & mAVE$\downarrow$ & mAAE$\downarrow$ \\

\hline

$\text{MV2D\cite{MV2D} }$ & V2-99 &$640\times 1600$ & C &
0.463 & 0.514 & 0.542 & 0.247 & 0.403 & 0.857 & 0.127 \\

$\text{BEVDepth\cite{bevdepth} }$ & V2-99 &$640\times 1600$ & C &
0.503 & 0.600 & 0.445 & 0.245 & 0.378 & 0.320 & 0.126 \\

$\text{HoP\cite{hop} }$ & V2-99 &$640\times 1600$ & C &
0.528 & 0.612 & 0.491 & 0.239 & 0.332 & 0.343 & \textbf{0.109} \\

$\text{SOLOFusion\cite{solofusion} }$ & ConvNeXt-B &$640\times 1600$ & C & 0.540 & 0.619 & 0.453 & 0.257 & 0.376 & 0.276 & 0.148 \\

$\text{BEVformer v2\cite{bevformerv2} }$ & InternImage-XL &$640\times 1600$ & C & 0.556 & 0.634 & 0.456 & 0.248 & \textbf{0.317} & 0.293 & 0.123 \\

$\text{StreamPETR\cite{StreamPETR}}$ & V2-99 &$640\times 1600$ & C &
0.550 & 0.636 & 0.479 & 0.239 & \textbf{0.317} & \textbf{0.241} & 0.119 \\

$\text{Sparse4D v2\cite{sparse4dv2}}^{\dagger}$ & V2-99 &$640\times 1600$ & C &
0.557 & 0.638 & 0.462 & 0.238 & 0.328 & 0.264 & 0.115 \\

$\text{SparseBEV\cite{sparsebev}}$ & V2-99 &$640\times 1600$ & C & 0.556 & 0.636 & 0.485 & 0.244 & 0.332 & 0.246 & 0.117 \\[0.25ex]

$\text{CRN\cite{crn}}$ & V2-99 &$640\times 1600$ & C+R & \textbf{0.575} & 0.624 & 0.416 & 0.264 & 0.456 & 0.365 & 0.130 \\[0.25ex]

$\text{RCBEVDet\cite{rcbevdet}}$ & V2-99 &$640\times 1600$ & C+R & 0.550 & 0.639 & \textbf{0.390} & \textbf{0.234} & 0.362 & 0.259 & 0.113 \\[0.25ex]

$\text{GeoBEV\cite{geobev}}$ & V2-99 &$640\times 1600$ & C & 0.543 & 0.635 & 0.409 & \textbf{0.234} & \textbf{0.317} & 0.284 & 0.122 \\[0.25ex]

$\text{HV-BEV)\cite{hvbev}}$ & V2-99 &$900\times 1600$ & C & 0.505 & 0.598 & 0.544 & 0.249 & 0.353 & 0.318 & 0.117 \\[0.25ex]

$\text{CAM3DNet(Ours)}$ & V2-99 &$640\times 1600$ & C & 0.5709 & \textbf{0.6413} & 0.4575 & 0.2419 & 0.3462 & 0.2764 & 0.1201 \\


\hline






\end{tabular}
}
\end{table*}

\subsection{Datasets and Metrics}

\textbf{nuScenes Dataset.} The nuScenes dataset \cite{nuscenes} comprises 1,000 driving scenes, partitioned into 700, 150, and 150 scenes for the training, validation, and test sets, respectively. Each scene spans 20 seconds and is keyframe-sampled at a frequency of 2 Hz. This configuration yields approximately 1.4 million camera images and 1.4 million accurately annotated 3D bounding boxes across 10 object categories (e.g., pedestrians, cars, and trucks). For evaluation, we adopt the official metrics: mean Average Precision (mAP) and the nuScenes Detection Score (NDS). The NDS is a consolidated metric derived from mAP and a set of True Positive (TP) error metrics, which specifically include Average Translation Error (ATE), Average Scale Error (ASE), Average Orientation Error (AOE), Average Velocity Error (AVE), and Average Attribute Error (AAE).

\textbf{Waymo Open Dataset.} The Waymo Open Dataset (WOD) \cite{waymo} provides camera data covering a horizontal field of view of $230^\circ$, with ground-truth 3D bounding boxes annotated up to a maximum range of 75 meters. Following standard evaluation protocols for camera-only 3D detection, we report the Longitudinal Error-Tolerant (LET) metrics: mLET-AP, mLET-AP-H, and mLET-APL. To ensure rigorous fairness and comparability, we strictly adhere to the standard training data split utilized by existing mainstream methods \cite{StreamPETR,petrv2,mvfcos3d,petr}.

\textbf{Argoverse 2 Dataset.} Similar to nuScenes, the Argoverse 2 (AV2) dataset \cite{argoverse} encompasses 1,000 scenes, which are divided into 700, 150, and 150 scenes for the training, validation, and test splits, respectively. Visual data is captured via seven ring cameras providing a cohesive $360^\circ$ field of view, supporting a perception range of up to 150 meters. The official evaluation metrics include mAP, standard TP error metrics (i.e., ATE, ASE, and AOE), and the Composite Detection Score (CDS), a unified metric uniquely tailored for the AV2 benchmark.

\begin{table*}[t]
\centering
\caption{Comparison on the Waymo open dataset validation split. 'C' indicates that the input data comes from the camera.
\label{tab:mwaymo dataset}}
\resizebox{\textwidth}{!}{
\begin{tabular}{c|cc|c|ccc}
\hline
\centering Methods & Backbone & Image Size & Input & mLETAP$\uparrow$ & mLETAPL$\uparrow$ & mLETAPH$\uparrow$ \\
\hline
MV-FCOS3D++\cite{mvfcos3d} & ResNet101-DCN & $1248\times 832$ & C & 0.522 & 0.379 & 0.484 \\
PETR-DN\cite{petr} & ResNet101 & $1248\times 832$ & C & 0.502 & 0.358 & 0.462  \\
PETRv2\cite{petrv2} & ResNet101 & $1248\times 832$ & C & 0.519 & 0.366 & 0.479  \\
StreamPETR\cite{StreamPETR} & ResNet101 & $1248\times 832$ & C & 0.553 & 0.399 & 0.517  \\
DenseBEV\cite{densebev} & ResNet101-DCN & $1248\times 832$ & C & \textbf{0.583} & {0.411} & \textbf{0.546}  \\
CAM3DNet(ours) & ResNet101 & $1248\times 832$ & C & {0.576} & \textbf{0.413} & {0.535}  \\
\hline
\end{tabular}
}
\end{table*}

\begin{table*}[t]
\centering
\caption{Comparison on the Argoverse 2 validation split. 'C' indicates that the input data comes from the camera.
\label{tab:argoverse dataset}}
\resizebox{\textwidth}{!}{
\begin{tabular}{c|cc|c|cc|ccc}
\hline
\centering Methods & Backbone & Image Size & Input & mAP$\uparrow$ & CDS$\uparrow$ &mATE$\downarrow$ & mASE$\downarrow$ & mAOE$\downarrow$ \\
\hline
$\text{BEVStereo\cite{bevstereo} }$ & V2-99 &$640\times 960$ & C & 0.146 & 0.104 & 0.847 & 0.397 & 0.901 \\

$\text{SOLOFusion\cite{solofusion} }$ & V2-99 &$640\times 960$ & C & 0.149 & 0.106 & 0.934 & 0.425 & 0.779 \\

$\text{Sparse4D v2\cite{sparse4dv2} }$ & V2-99 &$640\times 960$ & C & 0.189 & 0.134 & 0.832 & 0.343 & 0.723 \\

$\text{StreamPETR\cite{StreamPETR} }$ & V2-99 &$640\times 960$ & C & 0.203 & 0.146 & 0.843 & 0.321 & 0.650 \\

$\text{Far3D\cite{far3d} }$ & V2-99 &$640\times 960$ & C & 0.244 & 0.181 & {0.796} & 0.304 & \textbf{0.538}\\

$\text{DualViewDistill\cite{DualViewDistill} }$ & V2-99 &$640\times 960$ & C & {0.264} & \textbf{0.198} & \textbf{0.734} & 0.299 & 0.573\\

$\text{CAM3DNet(Ours)}$ & V2-99 &$640\times 960$ & C & \textbf{0.266} & {0.193} & 0.812 & \textbf{0.275} & 0.556\\
\hline
\end{tabular}
}
\end{table*}




\begin{table}[t]
\centering
\rev{
\caption{Repeated-run results of CAM3DNet on the nuScenes validation set under the ResNet-50 configuration. All runs use identical training settings and the checkpoint from the final epoch.}
\label{tab:seed_results}
\begin{tabular}{c|cc}
\hline
Seed & mAP (\%) $\uparrow$ & NDS (\%) $\uparrow$ \\
\hline
0 & 45.98 & 55.11 \\
1 & 45.74 & 54.52 \\
2 & 45.70 & 54.68 \\
\hline
Mean $\pm$ Std. & $45.81 \pm 0.15$ & $54.77 \pm 0.31$ \\
\hline
\end{tabular}
}
\end{table}
\subsection{Main Results}

\textbf{Results on the nuScenes Dataset.} Table~\ref{tab:nusc_val} presents a comprehensive evaluation of our CAM3DNet on the nuScenes dataset. \rev{Our method achieves strong performance across different backbone configurations. Under the ResNet-50 setting, CAM3DNet achieves 45.81 ± 0.15\% mAP and 54.77 ± 0.31\% NDS over three independent runs, indicating stable performance across random seeds. Its mAP remains the highest among the compared ResNet-50 methods. Compared with Far3D, CAM3DNet improves mAP and NDS by 1.40 and 0.70 percentage points, respectively.} Furthermore, this strong representation capability generalizes effectively to the official nuScenes test server, where CAM3DNet attains a highly competitive 57.09\% mAP and 64.13\% NDS.

\begin{figure}[t]
    \centering\includegraphics[width=0.75\linewidth]{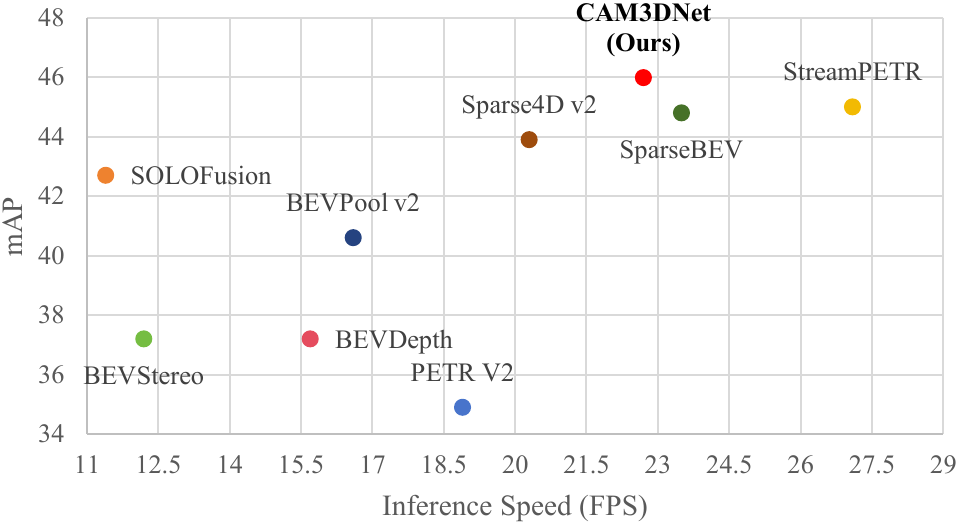}
    \caption{\textbf{The speed-accuracy performance of different models on nuScenes val set}.  All the methods use ResNet50 as the image backbone and the input size is set to $256\times704$. FPS is measured on a single RTX 3090 GPU using PyTorch with fp32 precision.}
    \label{fig:fps_figure}
\end{figure}
\begin{figure*}[t]
    \centering
    \includegraphics[width=\linewidth]{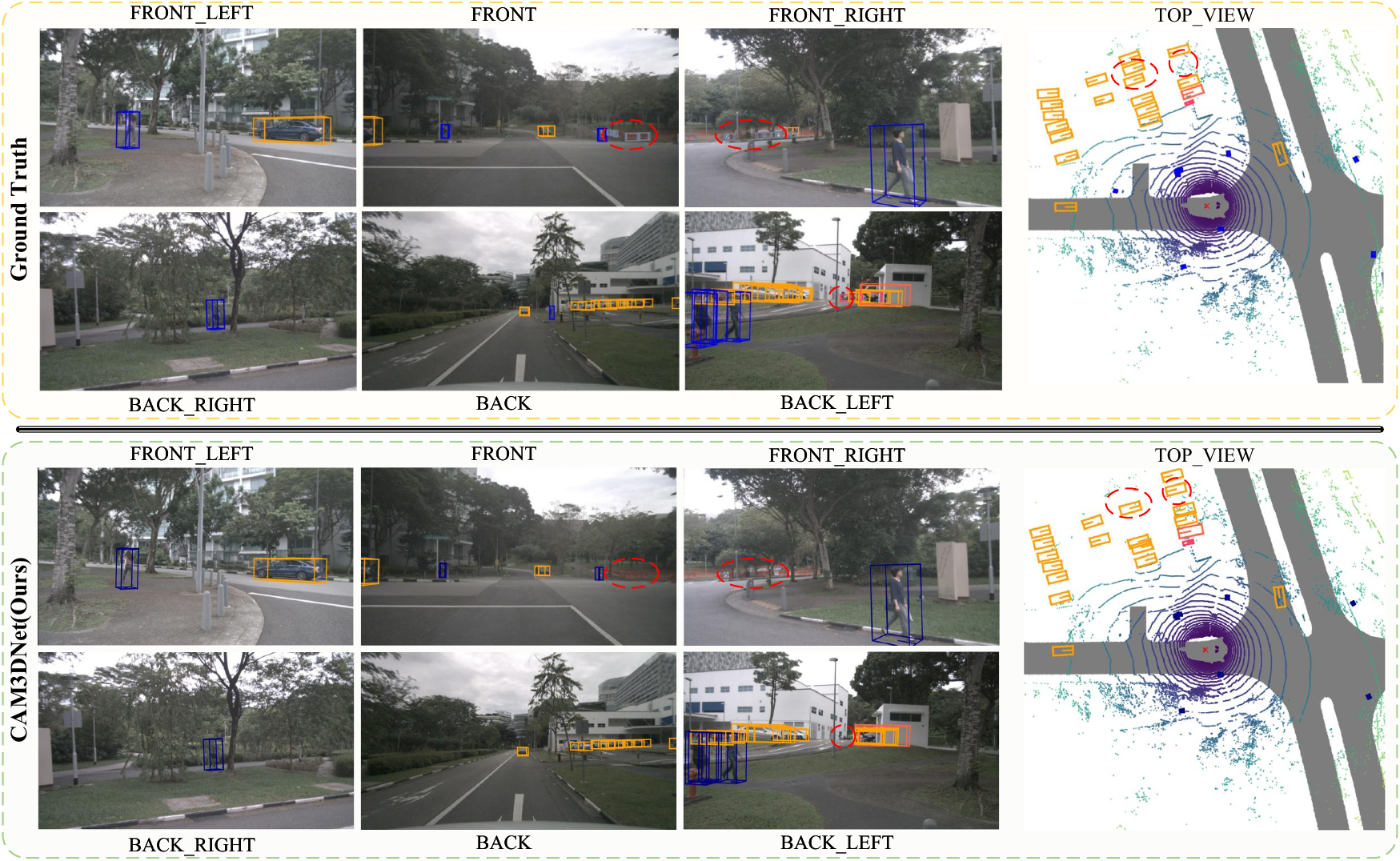}
    \caption{\textbf{Visualization results of CAM3DNet.} In the same scene, the ground truth (top) and prediction (bottom), along with their corresponding BEV plane (right), are visualized separately in two subplots. The failure cases in the image plane and BEV plane are marked with red circles.}
    \label{fig:visual result}
\end{figure*}
\begin{figure*}[t]
    \centering
    \includegraphics[width=\linewidth]{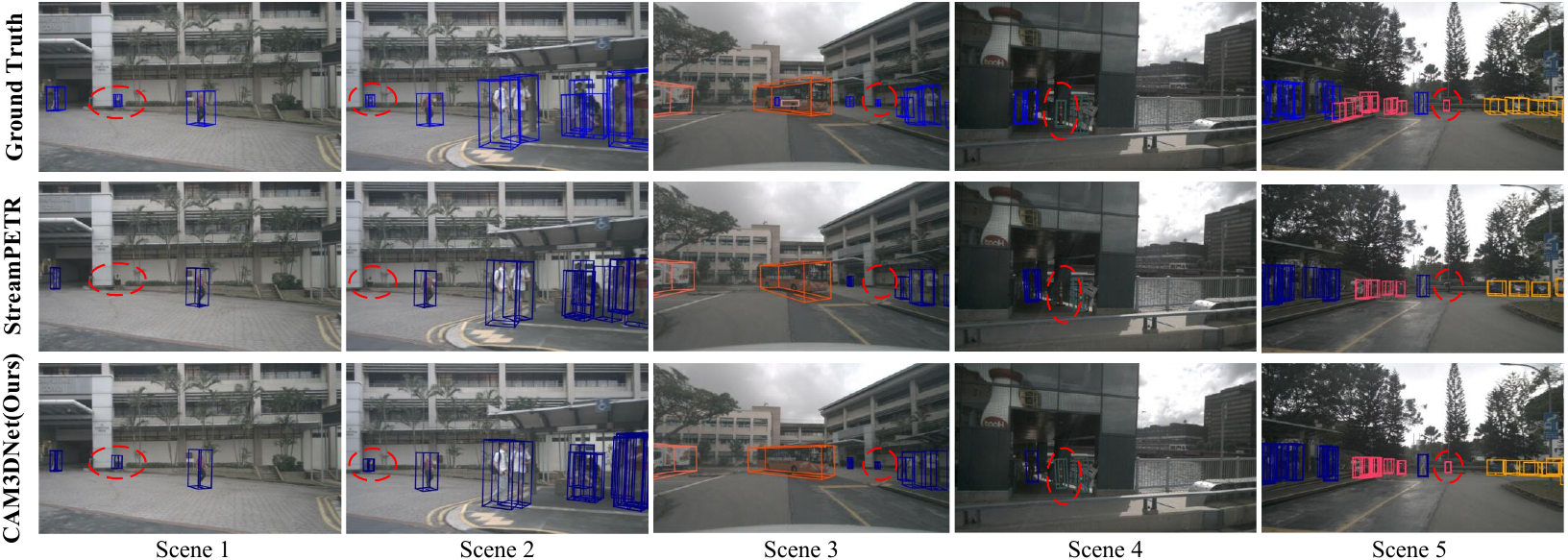}
    \caption{\textbf{Qualitative comparison between StreamPETR and CAM3DNet on the NuScenes dataset.} Red circles highlight representative regions where CAM3DNet produces more accurate and robust 3D detection results than StreamPETR.
    The failure cases in the image plane and BEV plane are marked with red circles.}
    \label{fig:detail result}
\end{figure*}

Beyond quantitative metrics, CAM3DNet demonstrates a highly favorable accuracy-efficiency trade-off, which is crucial for real-world autonomous driving deployments (Fig.~\ref{fig:fps_figure}). Specifically, under the ResNet-50 configuration, the integration of the proposed ASA and MSHS modules successfully mitigates redundant dense feature computations, effectively reducing the computational cost by 9.61 GFLOPs compared to the Far3D baseline. While our method introduces a marginal overhead of 8.56 M parameters and 84.17 GFLOPs relative to StreamPETR, this is well-justified by the substantial improvements yielded in small-object detection capabilities (Fig.~\ref{fig:detail result}). Finally, qualitative visualizations in challenging driving scenarios (Fig.~\ref{fig:visual result}) further corroborate CAM3DNet's robust perception of safety-critical targets, such as pedestrians and vehicles.

\textbf{Results on the Waymo Open Dataset.} To validate the efficacy of our method in long-range multi-scale perception, we evaluate CAM3DNet on the spatially expansive Waymo Open Dataset (Table~\ref{tab:mwaymo dataset}). Equipped with the ResNet-101 backbone, our approach outperforms StreamPETR by margins of 2.3\%, 1.4\%, and 1.8\% across the mLET-AP, mLET-APL, and mLET-APH metrics, respectively. These substantial gains indicate that our multi-scale feature mining strategy effectively mitigates the severe feature degradation typically associated with distant objects. Furthermore, CAM3DNet achieves performance highly competitive with dense BEV approaches (e.g., DenseBEV~\cite{densebev}). This demonstrates that our sparse composite queries can successfully capture wide-range semantic context without incurring the prohibitive computational overhead inherent to dense grid representations.

\textbf{Results on the Argoverse 2 Dataset.} To further assess geometric and scale estimation accuracy, we benchmark CAM3DNet on the Argoverse 2 dataset (Table~\ref{tab:argoverse dataset}). Leveraging the VoVNetV2-99 (V2-99) backbone~\cite{dd3d}, our method surpasses the Far3D baseline by 2.2\% in mAP and 1.2\% in CDS. More notably, we observe a prominent 2.9\% reduction in the Average Scale Error (mASE). This significant decrease in error strongly substantiates that the proposed ASA mechanism successfully rectifies sub-optimal query-geometry modeling, thereby facilitating highly precise, scale-aware 3D predictions.



\begin{table}[t]
\centering
\caption{Ablation on main components of CAM3DNet. \rev{CQ, ASA, and MSHS denote composite query, adaptive self-attention, and multi-scale hybrid sampling, respectively. Disabling CQ restores the baseline query construction while retaining the ROIHead; disabling MSHS removes fixed sampling while retaining learnable sampling. Values in parentheses indicate absolute gains over the baseline in percentage points.}}
\label{tab:main_component}
\begin{tabular}{c|ccc|cc}
\hline
\# & CQ & ASA & MSHS & mAP[\%]$\uparrow$ & NDS[\%]$\uparrow$ \\
\hline
1 & - & - & - & 44.41 & 54.07 \\
\hline
2 & \checkmark & - & - & 44.50 & 54.11 \\
3 & - & \checkmark & - & 45.67 & 54.65 \\
4 & - & - & \checkmark & 44.53 & 54.42 \\
\hline
5 & \checkmark & \checkmark & - & 45.86 & 54.76 \\
6 & \checkmark & - & \checkmark & 44.63 & 54.46 \\
7 & - & \checkmark & \checkmark & 45.89 & 55.07 \\
\hline
8 & \checkmark & \checkmark & \checkmark & \textbf{45.98}(+1.57) & \textbf{55.11}(+1.04) \\
\hline
\end{tabular}
\end{table}

\subsection{Ablation Study}
\label{sec: analysis}

In this section, we conduct comprehensive ablation studies on the nuScenes validation set to validate the efficacy of the proposed components in CAM3DNet. Unless otherwise specified, all experiments utilize a ResNet-50 backbone initialized with nuImages pre-trained weights~\cite{nuscenes}, operating at a resolution of $256\times704$, and are trained for 24 epochs without the CBGS~\cite{cbgs} strategy.


\textbf{Effectiveness of Core Components.} \rev{Table~\ref{tab:main_component} presents the ablation of CQ, ASA, and MSHS on the Far3D~\cite{far3d} baseline. ASA provides the largest standalone gains of 1.26 and 0.58 percentage points in mAP and NDS, respectively, highlighting its central role in CAM3DNet. Adding MSHS to the ASA-equipped baseline further improves mAP/NDS by 0.22/0.42 points, while incorporating CQ into this configuration yields additional gains of 0.09/0.04 points. The complete model achieves 45.98\% mAP and 55.11\% NDS, exceeding the ASA-only variant by 0.31 and 0.46 points, respectively. These results show that the simplified ASA-only variant already achieves competitive accuracy, while CQ and MSHS provide modest complementary improvements.}

\begin{table}[t]
 \centering
\caption{Ablation on main Components of composite query. We add the Global query (GQ), Temporal query (TQ) and Adaptive query (AQ) in order.}\label{tab:composite}
\begin{tabular}{c|ccc|cc}
\hline
  \# & GQ & TQ & AQ & mAP[\%]$\uparrow$ & NDS[\%]$\uparrow$ \\
\hline
 1 & \checkmark & - & - & 44.41 & 54.07 \\
 2 & - & \checkmark & - & 44.03 & 53.14 \\
 3 & - & - & \checkmark & 44.18 & 53.66 \\
 \hline
 4 & \checkmark & \checkmark & - & 44.90 & 54.81 \\
 5 & \checkmark & - & \checkmark & 45.07 & 54.89 \\
 6 & - & \checkmark & \checkmark & 44.73 & 54.47 \\
 \hline
 7 & \checkmark & \checkmark & \checkmark & \textbf{45.98} & \textbf{55.11} \\
\hline
\end{tabular}
\end{table}

\begin{figure*}[t]
    \centering

    \begin{minipage}[t]{0.49\linewidth}
        \centering
        \includegraphics[width=\linewidth]{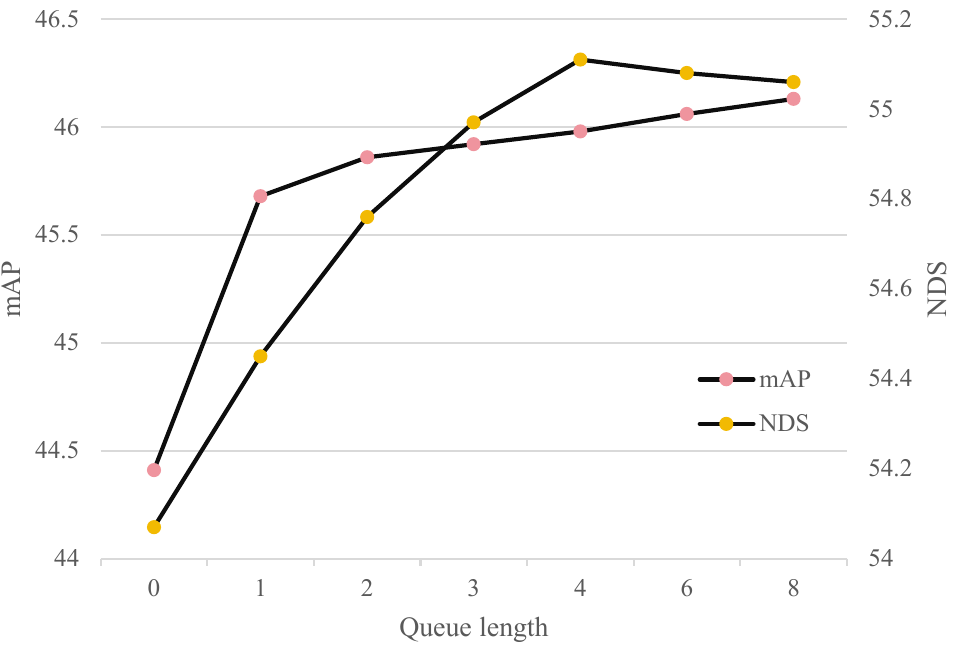}
    \end{minipage}
    \hfill
    \begin{minipage}[t]{0.49\linewidth}
        \centering
        \includegraphics[width=\linewidth]{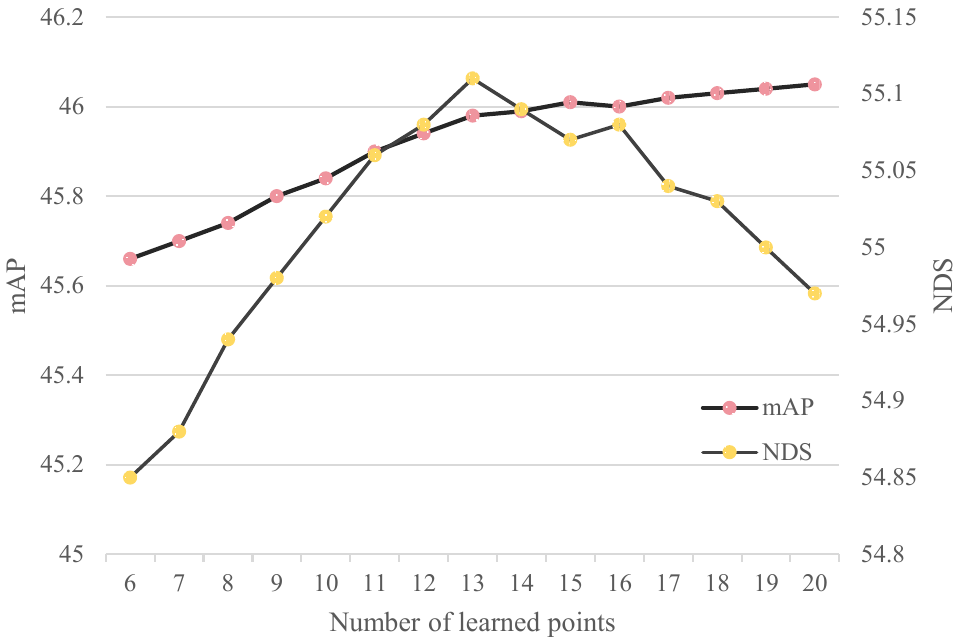}
    \end{minipage}

    \caption{\textbf{Comparison of (a) queue length for temporal queries and (b) the number of learnable points.} We investigate the impact of temporal information on model performance by varying the length of the input queue and the effect of varying the number of learnable points on model performance. The fixed points strategy was set to edge sample for the duration of the study.}
    \label{fig:queue length and learn points number}
\end{figure*}

\textbf{Ablation on Query Types.} Table~\ref{tab:composite} details the impact of Adaptive Queries (AQ), Temporal Queries (TQ), and Global Queries (GQ), utilizing the GQ-only setting as our fundamental baseline (row 1). We first investigate the temporal queue length (Fig.~\ref{fig:queue length and learn points number}(a)). While mAP steadily increases with longer queues, NDS peaks at 55.11\% at a queue length of 4. Adopting this optimal setting, the introduction of TQ yields improvements of 0.49\% mAP and 0.74\% NDS over the baseline (rows 1 vs. 4). Furthermore, integrating AQ into this configuration yields optimal performance, surpassing the GQ+TQ variant by 1.08\% mAP and 0.30\% NDS (rows 4 vs. 7). This substantial gain indicates that AQ effectively harnesses multi-scale features while providing robust priors for CQ. Finally, entirely replacing GQ with either TQ or AQ degrades mAP by 0.38\% and 0.23\% (rows 2 and 3), suggesting that GQ maintains a more balanced spatial distribution and quantity, thereby serving as a necessary foundation for achieving optimal performance.

\begin{table}[t]
\centering
\caption{Ablation on adaptive self-attention (ASA) with vanilla multi head self-attention (MHSA). ‘LN’ means linear-layer in SparseBEV \cite{sparsebev}, 'MLP' means two-layer Multi-Layer Perception in Sec.\ref{ssec:asa}.
\label{tab:asa}}
\begin{tabular}{c|c|cc}
\hline
Attention & Function & mAP[\%]$\uparrow$ & NDS[\%]$\uparrow$ \\
\hline
MHSA & - &  44.41 & 54.07 \\
\hline
\multirow{4}{*}{ASA(w/o LN)} & ${D^2}$ & \textbf{45.59} & \textbf{54.20} \\
 & $D$ & 44.76 & 53.86 \\
 & ${1}/{(1+D)}$ & 44.93 & 54.03 \\
\hline
\multirow{4}{*}{ASA(LN)} & ${D^2}/{2\varepsilon_i^2}$ & \textbf{46.07} & \textbf{54.65} \\
 & $D/{\varepsilon_i}$ & 45.06 & 53.76 \\
 & ${1}/{(1+{D/\varepsilon_i})}$ & 45.23 & 53.63 \\
\hline
\multirow{4}{*}{ASA(MLP)} & ${D^2}/{2\varepsilon_i^2}$ & \textbf{45.98} & \textbf{55.11} \\
 & $D/{\varepsilon_i}$ & 44.96 & 54.15 \\
 & ${1}/{(1+{D/\varepsilon_i})}$ & 45.35 & 54.67 \\
 
\hline
\end{tabular}
\end{table}

\textbf{Ablation on ASA Configurations.} Table~\ref{tab:asa} presents an evaluation of ASA compared to the vanilla Multi-Head Self-Attention (MHSA) and analyzes its internal architectural variants. Our optimal ASA design---utilizing a MLP and a distance function of $D^2/2\varepsilon_\mathrm{i}^2$---significantly outperforms vanilla MHSA, yielding substantial gains of 1.57\% mAP and 1.04\% NDS. We further ablate the scale estimation kernels. Compared to alternative strategies, such as normalized distance or Cauchy kernels, the $D^2/2\varepsilon_\mathrm{i}^2$ formulation achieves superior performance, indicating that it provides a more appropriate decay strategy for effectively modeling scale variations among queries. Finally, we compare weight generation strategies across MLP, Single Linear Network (LN), and non-linear (w/o LN) designs. While replacing LN with MLP incurs a negligible 0.09\% drop in mAP, it brings a notable 0.46\% improvement in the comprehensive NDS metric. This suggests that the deeper MLP structure possesses a stronger capacity for non-linear scale modulation.

\begin{table}[t]
\centering
\caption{Ablation on multi-scale hybrid sampling (MSHS). We incorporate learnable points (LS) and fixed-scale points (FS) into different generation method.\label{tab:mshs_1}}
\begin{tabular}{c|c|cc}
\hline
Method & Details & mAP$\uparrow$ & NDS$\uparrow$ \\
\hline
 Random & - & 45.03 & 53.63 \\
\hline
\multirow{3}{*}{Proposed} & LS + FS & \textbf{45.98} & \textbf{55.11} \\
 & LS + LN($wlh$) & 45.31 & 53.81 \\
 & LN(Concat(FC, $wlh$)) & 45.59 & 53.90 \\
\hline
\end{tabular}
\end{table}


\begin{table}[t]
\centering
\caption{Ablation on sampling point types and quantities. For configurations using fixed sampling points, the number of learnable points is fixed at 13. Conversely, when learnable sampling points are varied, the number of fixed points is fixed at 7.
\label{tab:mshs_2}}
\begin{tabular}{c|c|cc}
\hline
Type & Quantity & mAP[\%] $\uparrow$ & NDS[\%] $\uparrow$ \\
\hline
\multirow{3}{*}{Fixed points} & 1 & 45.86 & 54.76 \\
 & 7 & \textbf{45.98} & \textbf{55.11} \\
 & 9 & 45.92 & 55.02 \\
\hline
\multirow{3}{*}{Learnable points} & 6 & 45.66 & 54.85 \\
 & 13 & 45.98 & \textbf{55.11} \\
 & 20 & \textbf{46.05} & 54.97 \\
\hline
\end{tabular}
\end{table}

\textbf{Ablation on MSHS Strategies.} Table~\ref{tab:mshs_1} demonstrates that explicitly generating reference points from 3D bounding boxes significantly outperforms random sampling, achieving 45.98\% mAP and 55.11\% NDS. Furthermore, our fixed-scale strategy surpasses learning-based point generation, likely because the latter inherently lacks the explicit geometric priors required to model complex spatial structures. We then ablate the spatial layout and quantity of sampling points (Table~\ref{tab:mshs_2}, Fig.~\ref{fig:queue length and learn points number}(b))). Among fixed-point layouts, the 7-point edge sampling strategy yields the highest NDS, outperforming center and corner sampling by better capturing geometric structures under spatial occlusion or truncation. Regarding learnable points, NDS optimally peaks at a count of 13. Consequently, the combination of 7 fixed edge points and 13 learnable points serves as our final MSHS configuration for optimal multi-scale feature aggregation.

\begin{figure*}[t]
    \centering

    \begin{minipage}[t]{0.49\linewidth}
        \centering
        \includegraphics[width=\linewidth]{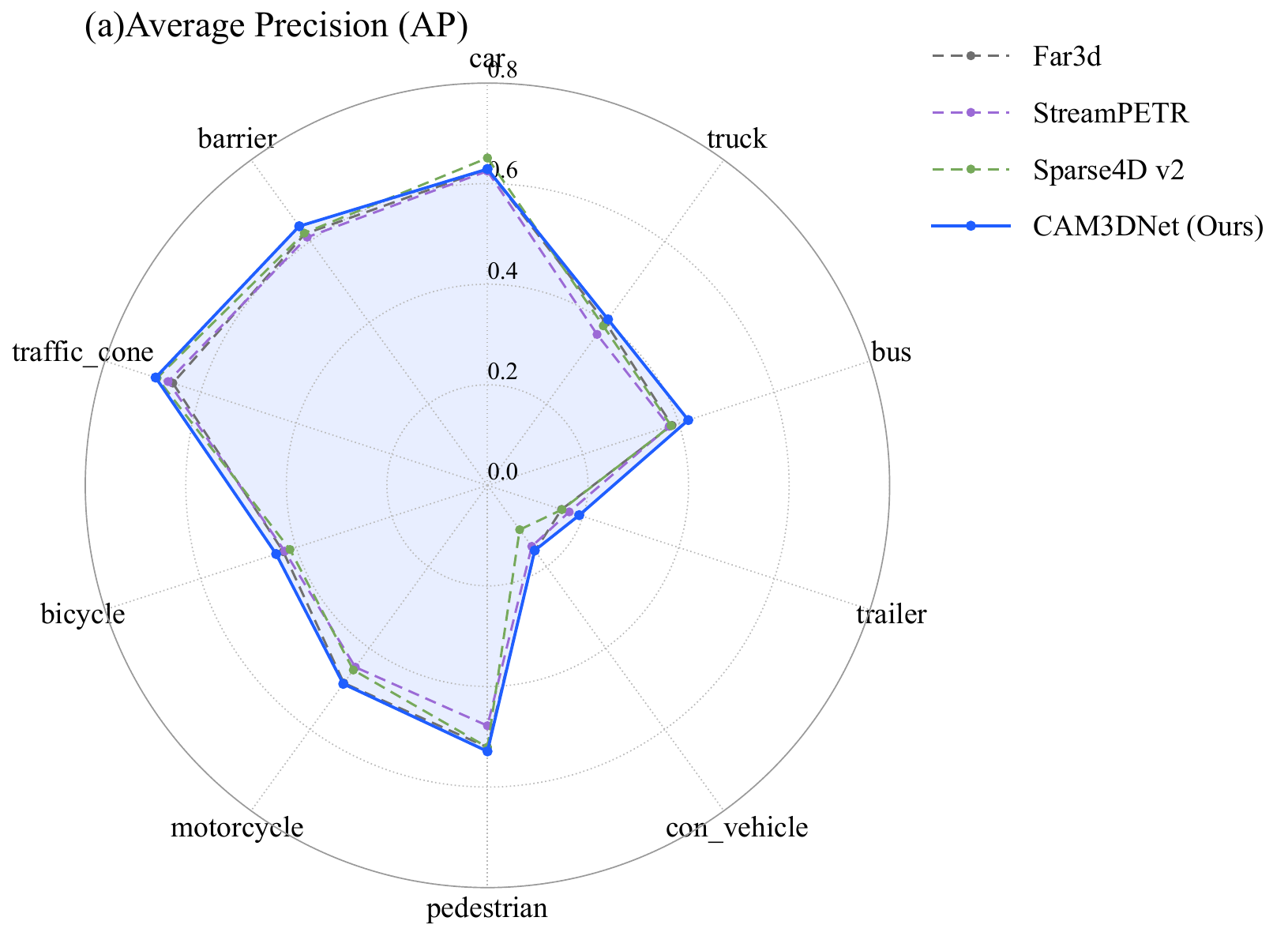}
    \end{minipage}
    \hfill
    \begin{minipage}[t]{0.49\linewidth}
        \centering
        \includegraphics[width=\linewidth]{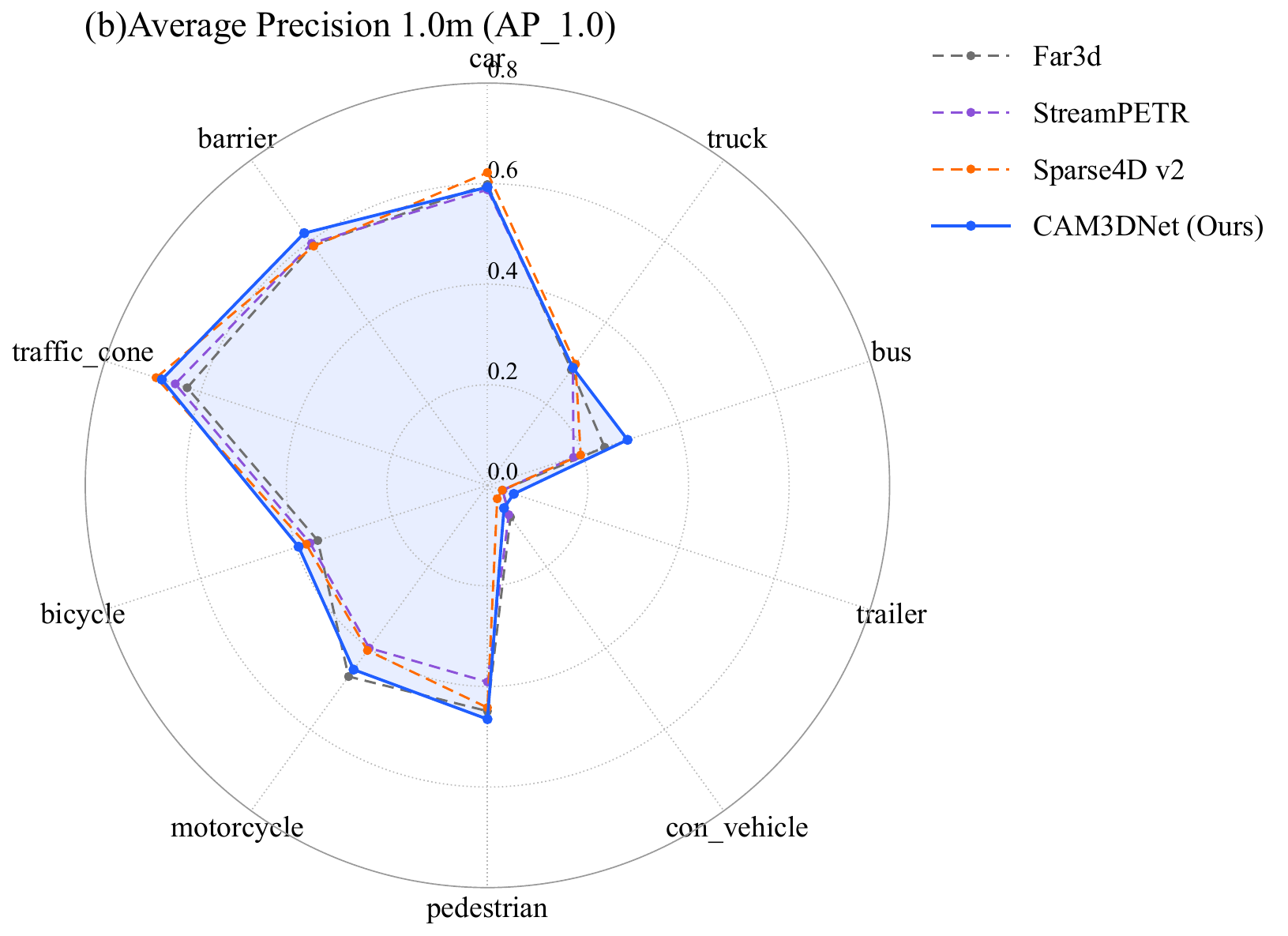}
    \end{minipage}

    \caption{\textbf{Comparison of (a) Average Precision (AP) and (b) Average Precision under 1.0m distance threshold (AP\_1.0) between CAM3DNet and SOTA methods.} On the nuScenes validation set, we compare the average precision of different methods across each object category. ``con\_vehicle'' denotes construction vehicle.}
    \label{fig:ap_class_1}
\end{figure*}


\textbf{Average Precision per Category.}
\rev{
Figure~\ref{fig:ap_class_1} presents the official class-wise AP(Fig.~\ref{fig:ap_class_1}(a))) and
AP$_{1.0}$(Fig.~\ref{fig:ap_class_1}(b))) results on the nuScenes validation set. To specifically
examine representative small-object categories, we compare pedestrian,
bicycle, motorcycle, and traffic cone. Compared with StreamPETR,
CAM3DNet improves the AP of pedestrian, motorcycle, and traffic cone
by 2.36, 4.05, and 4.10 percentage points, respectively, while the AP
of bicycle decreases by 2.03 points. The arithmetic mean over these
four categories increases from 51.60\% to 53.72\%. Under the stricter
1.0~m matching threshold, CAM3DNet also achieves gains of 4.39, 7.17,
and 5.75 points for pedestrian, motorcycle, and traffic cone,
respectively. These results indicate improved detection and localization
for most of the representative small-object categories, although the
benefit is not uniform across all classes.
}

\begin{table}[t]
\centering
\rev{
\caption{Distance-stratified detection performance on the nuScenes
validation set. Results are averaged over five vehicle categories
(car, truck, bus, trailer, and construction vehicle), which share
the official maximum detection range of 50~m. The mAP is computed
using the nuScenes center-distance matching thresholds of 0.5, 1.0,
2.0, and 4.0~m. Values in parentheses denote absolute differences
from StreamPETR in percentage points.}
\label{tab:distance_eval}
\begin{tabular}{l|ccc}
\hline
Method
& 0--20~m
& 20--35~m
& 35--50~m \\
\hline
StreamPETR~\cite{StreamPETR}
& 56.20
& \textbf{34.61}
& 13.08 \\
CAM3DNet
& \textbf{56.74} $(+0.54)$
& 32.79 $(-1.82)$
& \textbf{14.06} $(+0.98)$ \\
\hline
\end{tabular}
}
\end{table}

\textbf{Distance-Stratified Evaluation.}
\rev{
We further evaluate detection performance over the 0--20~m, 20--35~m, and 35--50~m ranges. As shown in Table~\ref{tab:distance_eval}, CAM3DNet improves mAP by 0.54 and 0.98 percentage points in the 0--20~m and 35--50~m ranges, respectively. In particular, the improvement in the most challenging 35--50~m range provides targeted evidence of enhanced long-range detection. However, CAM3DNet obtains a 1.82-point lower mAP in the 20--35~m range, indicating that the improvement is concentrated in the near and long-range regimes rather than being uniform across all distances.
}

\begin{table*}[t]
\centering
\caption{Ablation on FLOPs and Parameters. We analyze the parameter count and computational complexity of the proposed CAM3DNet in detail, including the contribution of each module.}
\label{tab:flops and params}
\resizebox{\textwidth}{!}{
\begin{tabular}{c|c|cc|cc}
\hline
\centering Module & Input Size & FLOPs(G) & FLOPS(\%) & Params(M) &Params(\%) \\
\hline

 ResNet50 & $(6\times3\times256\times704)$ & 88.785 & 43.717 & 23.455 & 54.550  \\
 \hline
 FPN & \makecell[c]{$(6\times256\times64\times176)$,$(6\times512\times32\times88)$,\\$(6\times1024\times16\times44)$,$(6\times2048\times8\times22)$} & 17.123 & 8.817 & 3.279 & 7.626  \\
 \hline
 ROIHead & \makecell[c]{$(6\times256\times32\times88)$,$(6\times256\times16\times44)$,\\$(6\times256\times8\times22)$,$(6\times256\times4\times11)$} & 73.274 & 37.73 & 10.657 & 24.785  \\
 \hline
 3DHead & \makecell[c]{$(6\times256\times32\times88)$,$(6\times256\times16\times44)$,\\$(6\times256\times8\times22)$,$(6\times256\times4\times11)$} & 15.036 & 7.742 & 5.606 & 13.038  \\
\hline
 Total & - & 194.217 & 100 & 42.997 & 100  \\
\hline
\end{tabular}
}
\end{table*}

\textbf{Analysis of Computational Complexity.} 
\rev{
As shown in Table~\ref{tab:nusc_val}, CAM3DNet maintains a parameter
budget comparable to Far3D (42.997 M vs. 43.004 M), while reducing
FLOPs from 203.825 G to 194.217 G under the ResNet-50 configuration.
Compared with StreamPETR, its higher computational cost is associated
with larger feature maps and the ROIHead.
Table~\ref{tab:flops and params} further shows that the ROIHead,
comprising a YOLOX head and a depth estimator, accounts for
37.73\% of the total FLOPs, second only to the backbone.
This component is shared with the Far3D baseline and retained across
all ablation variants, providing the 2D detection and depth priors
used to initialize adaptive queries. Its substantial computational
cost motivates the development of a lighter ROIHead for future
efficiency improvements.
}

\section{Conclusions}

In this paper, we have presented CAM3DNet, an efficient sparse query-based 3D object detection framework tailored for dynamic multi-scale feature mining in autonomous driving scenarios. By encoding comprehensive spatiotemporal cues within Composite Queries (CQ), CAM3DNet establishes a robust global representation, which is subsequently integrated via an Adaptive Self-Attention (ASA) mechanism to successfully model intricate query-geometry interactions. Furthermore, as spatial aggregation is explicitly guided by our Multi-Scale Hybrid Sampling (MSHS) strategy, the reference points seamlessly combine fixed and learnable representations, thereby ensuring precise detection performance even when localizing small objects in cluttered environments. To mitigate the inherent depth ambiguity of camera-only data, advanced low-light data augmentation and cross-sensor alignment are highlighted as promising directions for future optimization. Extensive experiments demonstrate that CAM3DNet achieves state-of-the-art accuracy and exhibits superior generalization capabilities across the Waymo and Argoverse benchmarks, underscoring its viability for robust real-world perception.

\section*{Acknowledgments}

This work was supported in part by the Major Scientific and Technological Innovation Project of Xianyang (L2023-ZDKJ-JSGG-GY-018 \& L2025-ZDKJ-ZDGG-RGZN-001); in part by the Natural Science Basic Research Program - General Project (Youth Project) of Shaanxi Province under grant 2025JC-YBQN-939.

We would like to thank Intelligent Equipment and Technology Research Division, Northwest Institute of Mechanical and Electrical Engineering, and the College of Intelligent Science, National University of Defense Technology for providing the research environment and computing resources. Additionally, we appreciate the open-source datasets used in this study, which have made 
a significant contribution to our research.

\section*{Declarations of Conflict of Interest}

The authors declared that they have no conflicts of interest to this work.

\noindent








\bibliography{cas-refs}

\begin{figure}[h]%
\centering
\includegraphics[width=0.3\textwidth]{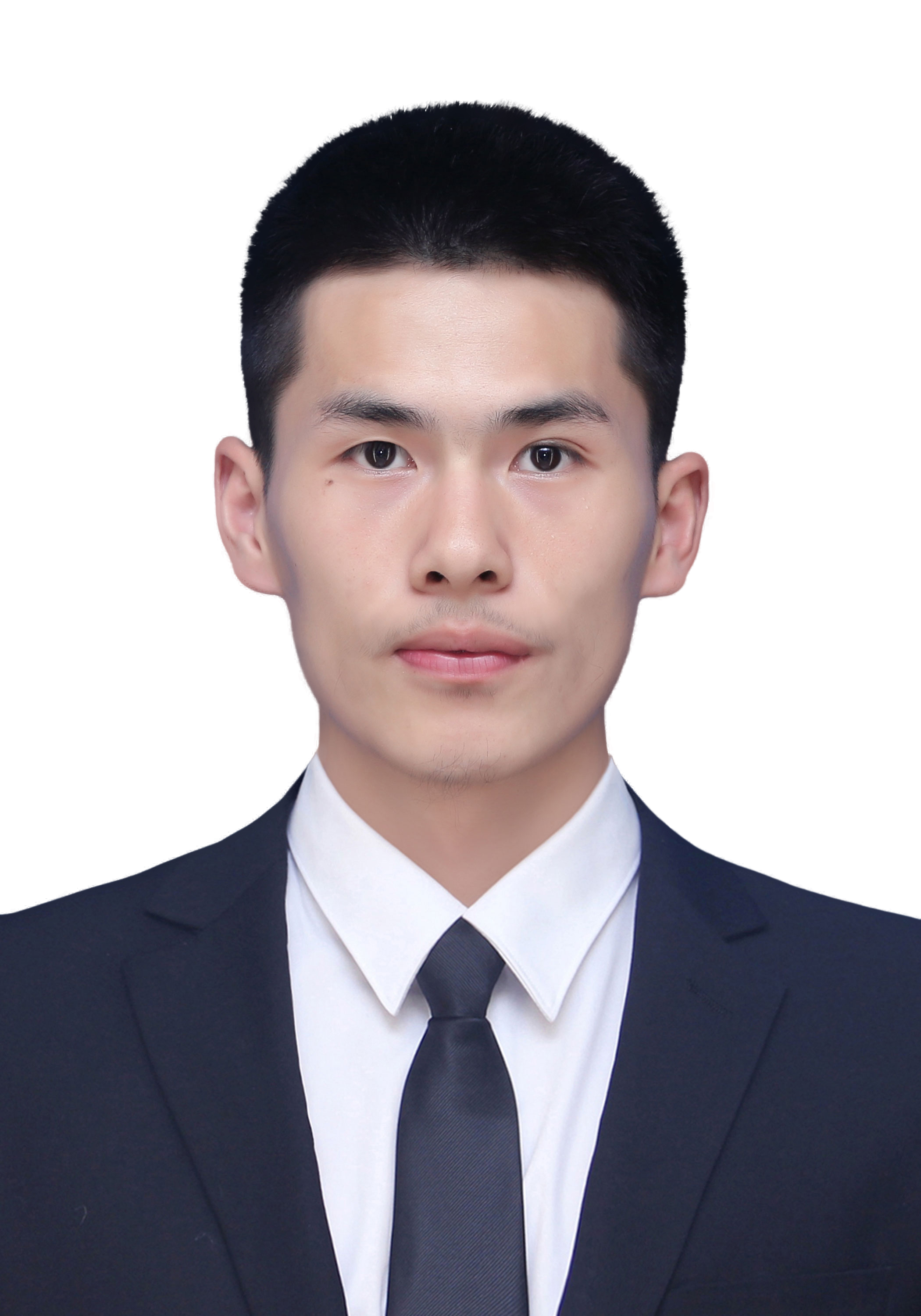}
\end{figure}

\noindent{\bf Mingxi Pang} received his B.E. degree from Nanjing University of Aeronautics and Astronautics, Nanjing, China, in 2022, and his master's degree from the Northwest Institute of Mechanical \& Electrical Engineering, Xianyang, China, in 2025. He is currently pursuing the Ph.D. degree at the National University of Defense Technology, Changsha, China.

His main research interests include machine learning, visual object detection, and multi-agent intelligence.

E-mail: mingxi\_pang@nudt.edu.cn



\begin{figure}[h]%
\centering
\includegraphics[width=0.3\textwidth]{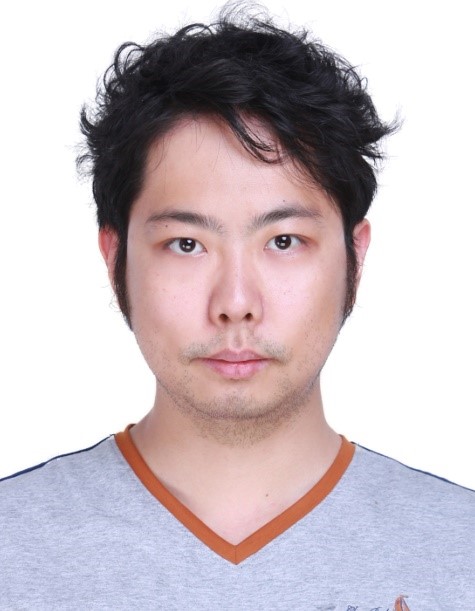}
\end{figure}

\noindent{\bf Dingheng Wang } received his B.E. and M.E. degrees from Xi'an Jiaotong University, Xi'an, China in 2010 and 2014, respectively. From 2014 to 2017, he worked as a software engineer at China Aerospace Science and Industry Corporation Limited. He completed his Ph.D. degree in Control Science and Engineering at Xi’an Jiaotong University in 2022. Currently, he is serving as an Associate Researcher at the Northwest Institute of Mechanical \& Electrical Engineering in Xianyang, Shaanxi, China. His research interests include neural network compression, efficient machine learning models, and intelligent systems.

E-mail: wangdai11@outlook.com (Corresponding author)



\begin{figure}[h]%
\centering
\includegraphics[width=0.3\textwidth]{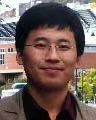}
\end{figure}

\noindent{\bf Jian Li } received the B.E. degree and the Ph.D. degree in control science and engineering from the National University of Defense Technology, Changsha, Hunan, China. He has been a Visiting Ph.D. Student with the Center for Intelligent Machines(CIM), McGill University, and a Visiting Scientist for cooperation research with the University of Hamburg. He is currently a Professor with the Department of Automation, College of intelligent science and technology, National University of Defense Technology, Changsha, China.

His research interests include computer vision, pattern recognition, image processing, and machine learning. He is one of the recipients received the 1st class Natural Science Award from Hunan Province, China, in 2018.

E-mail: lijian@nudt.edu.cn

\begin{figure}[h]%
\centering
\includegraphics[width=0.3\textwidth]{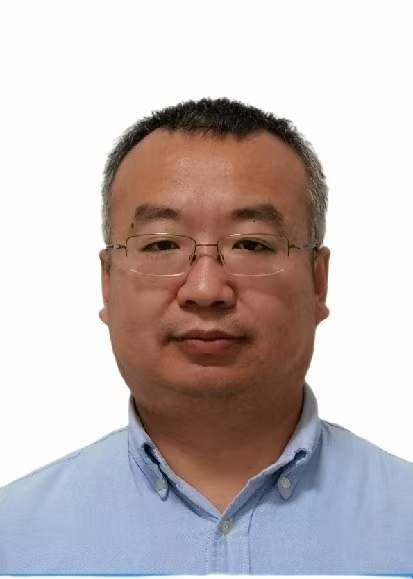}
\end{figure}

\noindent{\bf Zhenping Sun } received the Ph.D. degree in control science and engineering from the school of intelligent science and technology, National University of Defense Technology, Changsha, China, in 2004.

He is currently an Professor with the Department of Automation, College of intelligent science and technology, National University of Defense Technology, Changsha, China. His research interests include autonomous driving architecture, human-machine interaction, decision-making and planning technologies.

E-mail: sunzhenping@nudt.edu.cn

\begin{figure}[h]%
\centering
\includegraphics[width=0.3\textwidth]{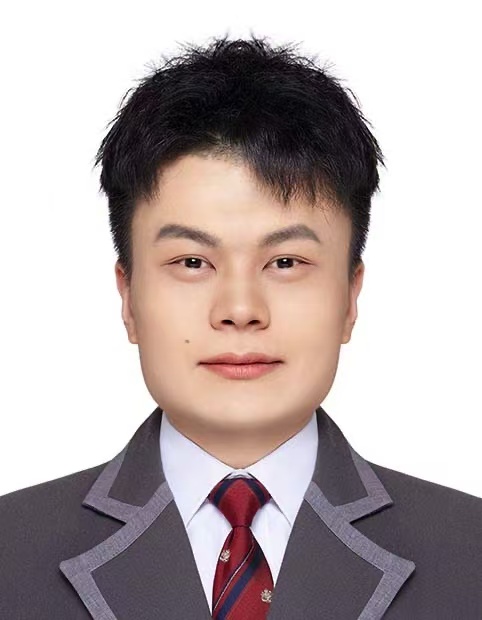}
\end{figure}

\noindent{\bf Zekun Li } received the B.S. degree from Xi'an University of Posts and Telecommunications,China,in 2018. He is currently pursuing the Ph.D.degree with the Department of Instrument Science and Technology, Xidian University,
Xi'an, China. Currently, he is serving as an Associate Researcher at the Northwest Institute of Mechanical \& Electrical Engineering in Xianyang, Shaanxi, China.
His research interests include neural computing, pattern recognition and computer vision.

E-mail: zkliberal@163.com

\begin{figure}[h]%
\centering
\includegraphics[width=0.3\textwidth]{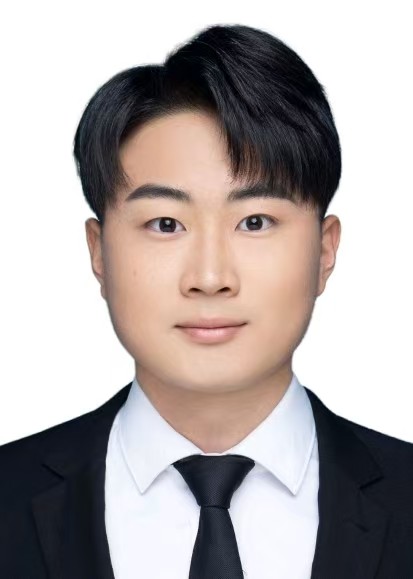}
\end{figure}

\noindent{\bf Zhihang Wang } received his B.E. degrees from Chang‘an University, Xi'an, China in 2020, and his master's degree from the Northwest Institute of Mechanical \& Electrical Engineering, Xianyang, Shaanxi, China, in 2025. He is currently an Associate Researcher at the same institute.

His main research interests include machine learning vision detection and pattern recognition.

E-mail: 1350152250@qq.com

\begin{figure}[h]%
\centering
\includegraphics[width=0.26\textwidth]{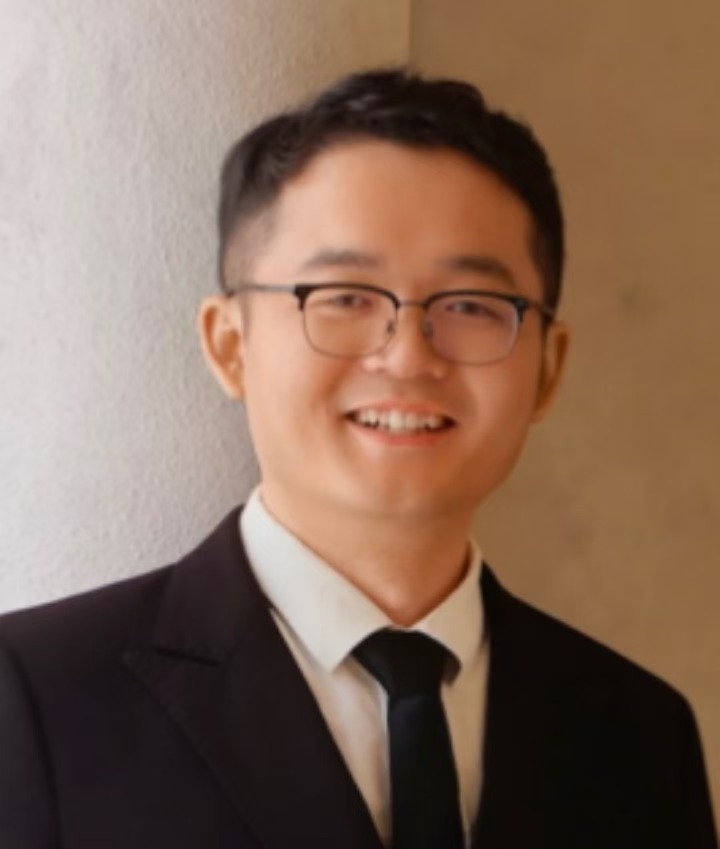}
\end{figure}

\noindent{\bf Yiqin Yang } received the Ph.D. degree in Control Science and Engineering from Tsinghua University, Beijing, China, in 2024. Currently, he is an assistant professor with the Institute of Automation, Chinese Academy of Science (CASIA). His current research interests include reinforcement learning, large language models and embodied intelligence. Dr. Yang serves as the Area Chair for ICLR 2026 and NeurIPS 2026.

E-mail: yiqin.yang@ia.ac.cn

\begin{figure}[h]%
\centering
\includegraphics[width=0.3\textwidth]{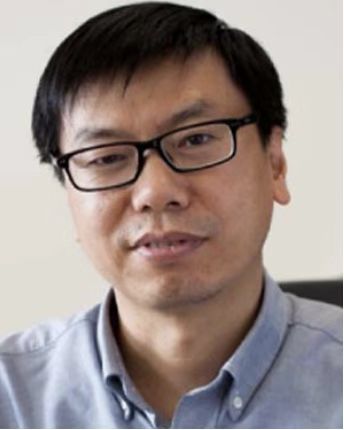}
\end{figure}

\noindent{\bf Bo Xu } received the BS degree from Zhejiang University, Hangzhou, China, in 1988, and the MS and PhD degrees from the University of Chinese Academy of Sciences, Beijing, China, in 1992 and 1997, respectively. Currently, he is a professor with the Institute of
Automation, Chinese Academy of Science(CASIA). And now is the director of CASIA, associate director of CAS(Chinese Academy of Sciences) Center for Excellence in Brain Science and Intelligence
Technology(CEBSIT). His research interests include spans the area of speech recognition, spoken dialogue, multimodality information processing, brain-inspired learning and decision intelligence. He once was the chairman of Special Interesting Group in International Association of Chinese Spoken language Processing(SIG-ISCSLP) between 2001-2006, and the Coordinator of International Consortium for Speech Translation Advanced Research (C-Star). He was steering committee member of National High-Tech Program in the fields of intelligence information processing and interactive interfaces from 2004-2010. He won several prizes including the best paper of ISCSLP, outstanding young scientist of CAS and outstanding achievement in national media technology development. He has published more
than 300 papers and holds 50 patents in the fields. He now serves as steering committee member of National Programme of New Generation Artificial Intelligence from 2017.

E-mail: xubo@ia.ac.cn

\end{document}